%% file: main.tex
\documentclass{article}

\usepackage[accepted]{icml2024}  

\usepackage[utf8]{inputenc} 
\usepackage[T1]{fontenc}    
\definecolor{mydarkblue}{rgb}{0,0.08,0.45} \usepackage[colorlinks=true, linkcolor=mydarkblue, urlcolor=blue, citecolor=mydarkblue, filecolor=mydarkblue]{hyperref}
\usepackage{url}            
\usepackage{booktabs}       
\usepackage{amsfonts}       
\usepackage{nicefrac}       
\usepackage{microtype}      
\usepackage{xcolor}         
\usepackage{color-edits}
\usepackage{graphicx}
\usepackage{amsmath}
\usepackage{subcaption}
\usepackage{wrapfig}

\addauthor{sm}{violet}
\addauthor{kr}{blue}
\addauthor{ga}{brown}

\begin{document}

\twocolumn[
\icmltitle{Interpretability in Action: Exploratory Analysis of VPT, a Minecraft Agent}
\icmlsetsymbol{equal}{*}

\begin{icmlauthorlist}
\icmlauthor{Karolis Jucys}{equal,bath}
\icmlauthor{George Adamopoulos}{equal,mila,mcgill,udem}
\icmlauthor{Mehrab Hamidi}{mila,mcgill}
\icmlauthor{Stephanie Milani}{cmu}
\icmlauthor{Mohammad Reza Samsami}{mila,udem}
\icmlauthor{Artem Zholus}{mila,poly}
\icmlauthor{Sonia Joseph}{mila,mcgill}
\icmlauthor{Blake Richards}{mila,mcgill,cifar}
\icmlauthor{Irina Rish}{mila,udem,cifar}
\icmlauthor{{\"O}zg{\"u}r~{\c{S}}im{\c{s}}ek}{bath}
\end{icmlauthorlist}

\icmlaffiliation{bath}{University of Bath}
\icmlaffiliation{cmu}{Carnegie Mellon University}
\icmlaffiliation{mila}{Mila - Quebec AI Institute}
\icmlaffiliation{mcgill}{McGill University}
\icmlaffiliation{udem}{Université de Montréal}
\icmlaffiliation{poly}{École Polytechnique de Montréal}
\icmlaffiliation{cifar}{CIFAR}

\icmlcorrespondingauthor{Karolis Jucys}{kr711@bath.ac.uk}

\vskip 0.3in
]

\printAffiliationsAndNotice{\icmlEqualContribution} 

\begin{abstract}
Understanding the mechanisms behind decisions taken by large foundation models in sequential decision making tasks is critical to ensuring that such systems operate transparently and safely.
In this work, we perform exploratory analysis on the Video PreTraining (VPT) Minecraft playing agent, one of the largest open-source vision-based agents.
We aim to illuminate its reasoning mechanisms by applying various interpretability techniques.
First, we analyze the attention mechanism while the agent solves its training task---crafting a diamond pickaxe.
The agent pays attention to the last four frames and several key-frames further back in its six-second memory.
This is a possible mechanism for maintaining coherence in a task that takes 3--10 minutes, despite the short memory span.
Secondly, we perform various interventions, which help us uncover a worrying case of goal misgeneralization: VPT mistakenly identifies a villager wearing brown clothes as a tree trunk when the villager is positioned stationary under green tree leaves, and punches it to death.\footnote{Videos, code and more: \href{https://sites.google.com/view/vpt-mi/}{https://sites.google.com/view/vpt-mi/}}

\end{abstract}

\input{sections/10_introduction}
\input{sections/20_related_work}
\input{sections/30_background}
\input{sections/40_visualisation}
\input{sections/50_intervention}
\input{sections/60_discussion_and_limitations}
\input{sections/90_conclusion}

\section*{Acknowledgements}

This work was supported by the UKRI Centre for Doctoral Training in Accountable, Responsible and Transparent AI (ART-AI) [EP/S023437/1], the University of Bath, the Canada CIFAR AI Chair Program [I.R.], the Canada Excellence Research Chairs Program [I.R.], NSERC (Discovery Grant: RGPIN-2020-05105; Discovery Accelerator Supplement: RGPAS-2020-00031; Arthur B. McDonald Fellowship: 566355-2022) [B.R.], and CIFAR (Canada AI Chair; Learning in Machine and Brains Fellowship) [B.R.].
This research was enabled in part by Calcul Québec and the Digital Research Alliance of Canada.
We thank Rachael Bedford, Daniel Beechey, Brandon Houghton, Gabija Juce, Anssi Kanervisto, and Rohin Shah for helpful comments and discussions.

\section*{Contributions}

KJ performed the experiments in sections 4--5 and wrote the initial drafts of sections 3--7 of the paper.
GA performed the experiments in appendix sections A, and C and wrote the initial drafts of sections 1--2 and appendix sections A, C.
MH performed the experiments in appendix section B and wrote the initial draft of the corresponding section.
MS, AZ, and SJ provided the initial code used for ablation experiments and gave support in using it.
SM, MS, AZ, SJ, BR, IR, and {\"O}{\c{S}} advised on the project and contributed to writing.

\bibliography{main}
\bibliographystyle{icml2024}

\newpage
\appendix
\onecolumn
\input{sections/100_appendix}

\end{document}

%% file: sections/10_introduction.tex
\section{Introduction}

Large transformer-based models have achieved significant success in autoregressively predicting the next token across various modalities, including language, images, and audio.
Recently, these models have started to be used as agents in online settings.
Given the advancements in large-scale AI-based agents interacting with real and simulated worlds \citep{raad2024scaling,figure2024ai}, it is crucial to understand their decision-making processes both in and out of distribution.
VPT \citep{baker2022video} is one of the first large-scale transformer-based agents (250 million parameters).
The techniques that have been developed in the field of mechanistic interpretability have the potential to be useful for better understanding VPT and other agents like it.
We chose VPT as the model organism for our studies because it is one of the largest and most capable open-source embodied agentic models that act in a 3D environment.
It has also been used as a backbone for developing other agents \citep{lifshitz2023steve1,milani2023towards}.
There are more advanced agents, such as SIMA \citep{raad2024scaling}; however, they are not openly available for study.

\begin{figure*}[t]
  \centering
  \includegraphics[width=\textwidth]{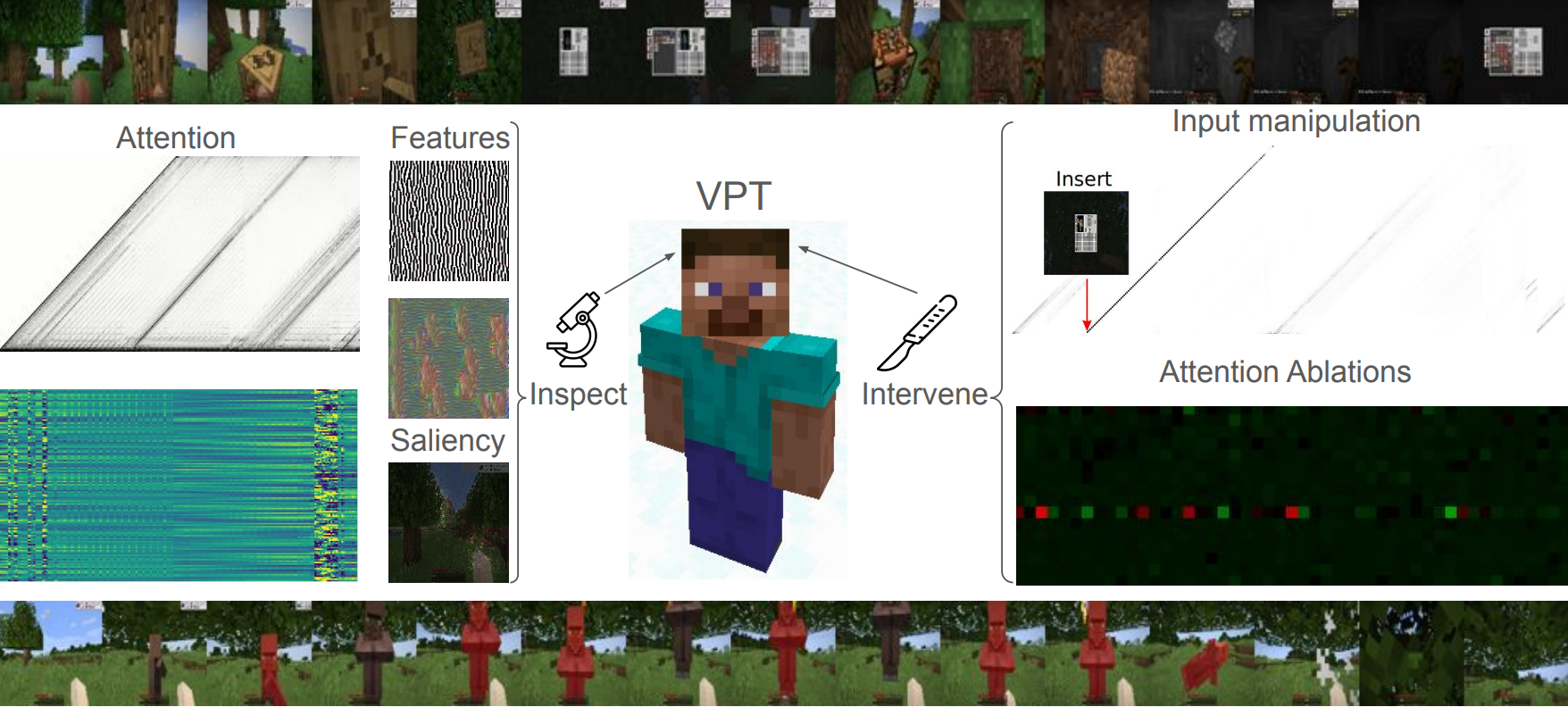}
  \caption{
  We use various interpretability techniques on the Minecraft playing agent VPT to better understand how it makes decisions.
  These include visualizing attention head weights and outputs, feature visualization, saliency maps, ablating attention head outputs, manipulating the input stream, and others.
  (Top) a part of a regular episode.
  (Bottom) an episode with a "villager-tree".
  See \href{https://youtu.be/g-jd6OyOcUs}{Video01}.}
  \label{fig:figure-1}
\end{figure*}

\paragraph{Mechanistic interpretability.} 
Mechanistic interpretability seeks to reverse engineer neural networks into the algorithms that the network weights have implemented throughout training.
The field has focused on circuit-based methods \citep{wang2022interpretability, elhage2021mathematical} and understanding attention heads \citep{mcdougall2023copy, olsson2022context}.
However, most of this work has been on large language models (LLMs), with limited investigation into other modalities.
We propose that applying mechanistic interpretability techniques that have been designed and tested on language to other modalities is essential, not only to validate the method but also to establish a foundation for the nuances of the various new modalities.
In addition, VPT has been fine-tuned with reinforcement learning (RL) and vision interpretability techniques often do not work for vision-based RL agents \citep{hilton2020understanding}.

\paragraph{Mechanistic interpretability on agents.}

Understanding the behavior of agentic models mechanistically poses more challenges than for traditional LLMs for several reasons.
First, isolating an agent's behavior requires replicating the environment and the actions leading to that behavior, which is difficult for agentic models.
For instance, \citet{hanna2023does} identified a circuit in GPT-2 that computes greater than by prompting the model with sentences like, "The war lasted from the year 1732 to the year 17" and analyzing its activations.
In contrast, for stochastic agentic models in dynamic environments, it is challenging to recreate slight variations of similar situations to assess if the model has generalized knowledge about a scenario.
While LLMs and their language outputs are stochastic, LLM interpretability researchers benefit from being able to carefully craft their prompts to elicit certain behaviors from the model.

Additionally, computing rollouts for large-scale agentic models is expensive, and comparing actions across different rollouts is complex.
Comparing action probabilities and analyzing rollouts proved most useful in assessing the effect of ablations on the overall behavior of VPT.
Rewards can sometimes provide a metric to compare the success of rollouts, but it does not capture the detail needed to understand if an agent's behavior has been altered and often can only show if the agent remains functional.
Finally, agentic models are trained to maximize long-term rewards, adding another layer of complexity to their decision-making processes compared to LLMs optimized for next-token prediction.

Our paper makes the following contributions (\autoref{fig:figure-1}):

\begin{itemize}
    \item We find clues on how the agent maintains coherence in a 3--10-minute task with only 6 seconds of memory.
    \item We show that single attention output ablations only influence actions when those actions are uncertain.
    \item We discover a new case of goal misgeneralization in the wild: when we place a brown villager under tree leaves and make it stand still, VPT mistakes the "villager-tree" for a real tree, begins to attack it, and often kills the villager (see \autoref{fig:villager-tree-example}, \href{https://youtu.be/VVkWWgwKf0M}{Video12}).
\end{itemize}

Goal misgeneralization is a failure mode where an agent competently pursues an unintended goal when in a novel situation \citep{shah2022goal,di2022goal}.
This incident further reinforces our initial motivation of trying to understand how large-scale agents interact with the world both in and out of distribution.

%% file: sections/20_related_work.tex
\section{Related work}

\paragraph{Mechanistic interpretability.}
Most mechanistic interpretability work has focused on reverse engineering circuits in LLMs \citep{wang2022interpretability, lieberum2023does, conmy2023towards}.
There has been some limited work on understanding transformer-based vision models.
For example, \citet{gandelsman2024interpreting} found that attention heads can have specialized roles, such as focusing on shapes, colors, and object counts.

\paragraph{Interpretable reinforcement learning.}
Reinforcement learning techniques can sometimes be substantially easier to interpret depending on what kind of model is used for the policy and value function.
Techniques such as linear function approximation, decision trees, and tabular methods often provide clear and understandable representations of the learned policies and value functions; however, almost all state-of-the-art RL uses deep RL \cite{milani2024explainable, glanois2021survey, kenny2023towards, dao2018deep}.

Recent advancements in interpretable RL include the development of frameworks and methodologies that aim to balance the performance and transparency of RL models.
For example, the SINDy-RL framework combines sparse dictionary learning with deep reinforcement learning to create interpretable and capable agents.
This approach leverages the Sparse Identification of Nonlinear Dynamics (SINDy) method to construct data-driven models that are not only effective but also offer insight into the underlying decision-making processes \citep{zolman2024sindyrl}.

Hierarchical reinforcement learning is another avenue that enhances interpretability.
This method splits tasks, where a higher-level controller decides on a sequence of subtasks while a lower-level controller handles specific actions within each subtask.
This hierarchical structure allows for a more intuitive understanding of the policy's behavior by breaking down complex decisions into simpler, more manageable parts \citep{barto2003hrl, xu2021interpretable}.

Additionally, explainable reinforcement learning techniques focus on post-hoc methods to explain the decisions of black-box models.
These include generating visualizations, using model-agnostic techniques like LIME \citep{ribeiro2016why}, Shapley values \citep{beechey2023explaining}, and developing algorithms that provide explanations for specific actions taken by the RL agent \citep{puiutta2020explainable}.

%% file: sections/30_background.tex
\section{Background}

\begin{figure*}[htbp]
  \centering
  \includegraphics[width=\textwidth]{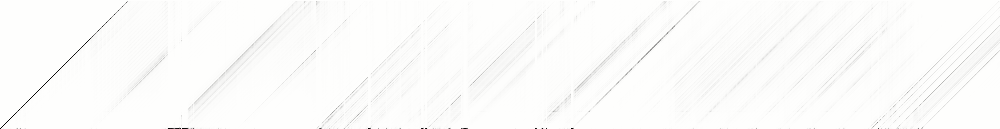}
  \includegraphics[width=\textwidth]{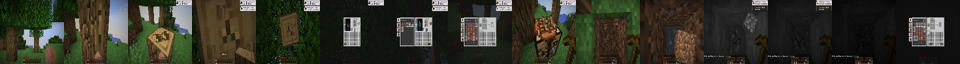}
  \includegraphics[width=\textwidth]{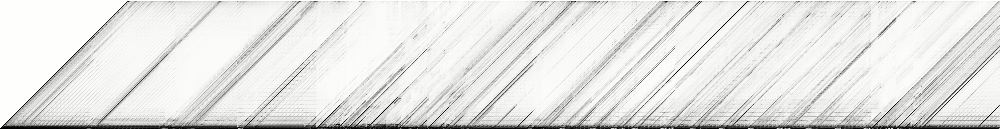}
  \caption{
  (Middle) Visualization of a trajectory up to crafting a stone pickaxe.
  The leftmost pixel of each frame corresponds to the time step in the attention plots.
  (Top) Attention weights of attention head 2.2---note the different pattern above the 3rd frame.
  This coincides with the camera moving up.
  The vertical axis is the 128 attention weights, the horizontal axis is time.
  (Bottom) Max attention weights over all attention heads.
  This shows that most attention is paid to 3--4 past frames and some key-frames.
  See \href{https://youtu.be/BeqSthHRyLA}{Video02} and \href{https://youtu.be/3GhhEysmSY4}{Video03}.}
  \label{fig:attention-over-time}
\end{figure*}

\paragraph{Environment---MineRL.}
VPT was trained in MineRL \citep{gussminerl2020}, which turns Minecraft into an environment.
Minecraft \cite{minecraft} is a 3D embodied sandbox video game known for its blocky graphics and open-world gameplay.
In survival mode, the player's goal is to gather resources, craft tools, and build shelters to survive against monsters. 
In MineRL, each rollout starts in a new procedurally generated world, with the agent acting in it from a first-person perspective.
The observations are the images of the game view.
The action space is the same as what a human would use---keyboard buttons and mouse.
One second of in-game time is equal to 20 time steps in the environment.
The tasks and rewards can be specified manually.
The most popular task has been obtaining a diamond.

\paragraph{Agent---VPT.}
VPT is a 250 million parameter model that was pre-trained using 70,000 hours of pseudo-labeled human Minecraft play videos and fine-tuned using reinforcement learning for over 230,000 in-game hours \citep{baker2022video}.
The architecture of the network is a combination of a residual convolutional neural network and a transformer.
The task the agent was fine-tuned on was obtaining a diamond pickaxe---one extra step beyond obtaining diamonds.

VPT's behavior is very consistent.
At the start of an episode, it goes towards the nearest tree, chops four logs, and proceeds to craft and mine its way through the subtasks (see \autoref{fig:curriculum}).
In about 80\% of episodes, it gets an iron pickaxe, the item required to mine diamonds.
It gets diamonds 20\% of the time and the diamond pickaxe 2\% of the time.

\paragraph{Agent---Steve-1.}
Steve-1 is a Minecraft agent that extends the capabilities of VPT by introducing instruction tuning, enabling it to follow natural language commands \citep{lifshitz2023steve1}.
It uses a richer dataset of human gameplay videos with text annotations, improving its understanding of complex behaviors.
This integration of visual and textual data during training gives it the ability to understand and act upon natural language instructions.
However, it can perform only short tasks that take several seconds.

\paragraph{Attention mechanism.}
In transformer models, \textit{attention weights} are computed by taking the softmax of the dot product of queries (Q) and keys (K), indicating how much focus each part of the input should receive.
\textit{Attention outputs} are then obtained by multiplying these weights by the values (V).
This allows the model to weigh the input elements appropriately when generating representations or predictions.
VPT has 4 attention layers with 16 heads in each.
We will use "head 2.3" to refer to layer 2 attention head 3.

%% file: sections/40_visualisation.tex
\section{Attention visualization}

We start our analysis by visualizing parts of the transformer attention mechanism.
VPT uses a transformer architecture with a context length of 128 frames (6.4 in-game seconds).
The agent takes between 3 to 10 minutes to make a diamond pickaxe from scratch when it succeeds at the task.
This involves a long sequence of subtasks, such as finding and chopping a tree, crafting different pickaxes, mining stone, finding and smelting iron, and others.
As a comparison, a proficient human takes on average 20 minutes to solve this task \citep{baker2022video}.
How does VPT "keep the thread" for so long?
That is, how does it know which part of the long sequential task it is currently solving?
We cannot help but notice the similarity to the famous musician Clive Wearing, who was struck by an extremely strong form of amnesia, which left him with about 7 seconds of memory.
Yet he could play long piano pieces \citep{sacks2007musicophilia}.

It could be that a single frame is enough to recognize what the agent has to do next.
For example, if it sees a stone pickaxe in the hotbar, but no iron pickaxe, it might be looking for iron ore.
However, the agent is quite robust to the items being placed in different slots of the hotbar, or even in the parts of inventory that are visible only when the inventory menu is open.
This would imply a more sophisticated mechanism, likely using its memory of past frames.
Given these constraints, it is likely that VPT has learned to play Minecraft in a different way than humans; something that we expect to be common among many large-scale AI agents.

\subsection{Attention weights}

Visualizing attention weights between tokens in LLMs has enabled the discovery of many interesting circuits, such as induction heads \citep{olsson2022incontext}.
We use a very similar technique in the sequential decision making domain in our work.
The main difference comes from the short context window of VPT---it does not have access to frames beyond the previous 128, while its task takes many thousands.

The top plot in \autoref{fig:attention-over-time} shows how much attention is being paid to each of the past 128 frames (vertical axis) over the first 1,000 frames of an episode (horizontal axis) by the attention head 2 on layer 2 (4 layers, 16 heads each).
The darker the color, the higher the attention weight, with 0\% represented by white and 100\% by black.
1,000 frames is 10--20\% of an average episode in which the agent obtains a diamond pickaxe.
The frames in the middle show the progression over that time---from chopping a tree to crafting a stone pickaxe.
The leftmost pixel of each frame corresponds to that exact moment in time on the above plot.
A diagonal line indicates that a specific frame is being attended to as the agent continues to interact with the environment.

\begin{figure*}[htbp]
  \centering
  \includegraphics[width=\textwidth]{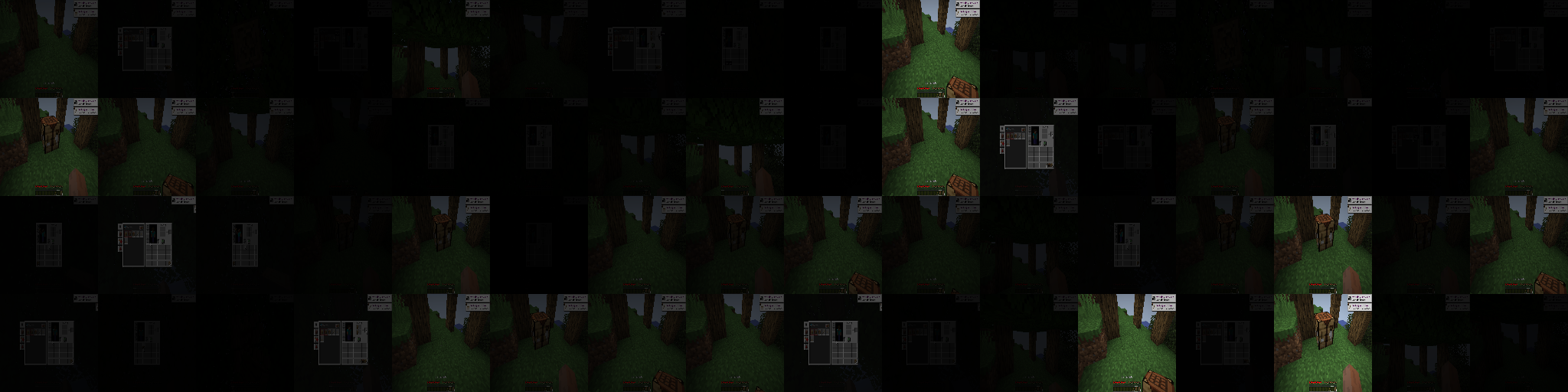}
  \caption{
  The frame that each attention head is paying the most attention to at a single point in time, right after placing a crafting table.
  Brightness indicates the magnitude of the attention.
  For example, heads 0.9 (1st row, 10th frame) and 1.9 are looking at the previous frame; heads 2.13, and 3.13 are looking at the current frame (crafting table placed).
  Other heads are looking at the inventory menu at different earlier times, some with the recipe book open, and some closed.
  See \href{https://youtu.be/3GhhEysmSY4}{Video03}.}
  \label{fig:4x16-top-attention-frame-418}
\end{figure*}

The bottom part of the top plot shows a short horizontal line (above the third visualized frame).
This indicates that the attention head is paying attention to the last frame it saw during the rollout.
This coincides perfectly with the agent looking up to chop the logs above it.
This does not happen when the agent looks down shortly after.
We can also observe how attention head 2.2 operates in two modes: one of paying attention to some specific frames in the past and the second of mainly looking at the last frame.
Many other attention heads show the same two-mode pattern (see \autoref{fig:all-heads-layer-0}--\autoref{fig:all-heads-layer-3} in the appendix).
The patterns are consistent between episodes.
Also, see \autoref{fig:4x16-top-attention-frame-418} for a visualization of attention weights at a single point in time.

To visualize what VPT pays the most attention to more clearly, we overlaid the attention weight patterns for all 64 attention heads by taking a maximum for each frame and each memory position (see the bottom plot of \autoref{fig:attention-over-time}).
We can see that the past 3--4 frames are always attended to, indicated by the thick dark line at the bottom; as well as some key-frames, seen as diagonal lines.
We surmise that the recent frames help describe its immediate past, while the key-frames allow it to recognize which part of the full task it is currently doing.
Some example key-frames include: a dirt block that was just destroyed (starting to dig down?), a cobblestone block that was just destroyed (dug further down?), and an achievement popup after making a stone pickaxe (can now mine iron?).
These are of course speculative and require more evidence.

\subsection{Attention outputs}

\begin{figure*}[htbp]
  \centering
  \begin{minipage}{.48\textwidth}
    \centering
    \includegraphics[width=\columnwidth]{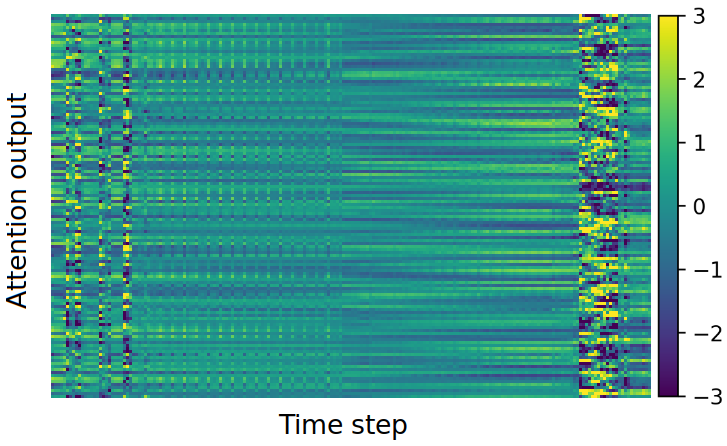}
    \caption{All 128 output z-scores of attention head 2.2 over the first 200 frames of a regular episode. The different pattern on the right coincides with the agent looking up. Regular vertical lines in the first half match the attacking arm looping every 4 frames. These disappear in the second half. The agent is still attacking, but the arm is replaced by a smaller object---an oak log it just chopped. See \href{https://youtu.be/TbTBWdb6jSo}{Video04}.}
    \label{fig:activation-viewer-2.2}
  \end{minipage}%
  \hfill
  \begin{minipage}{.48\textwidth}
    \centering
    \includegraphics[width=0.8\columnwidth]{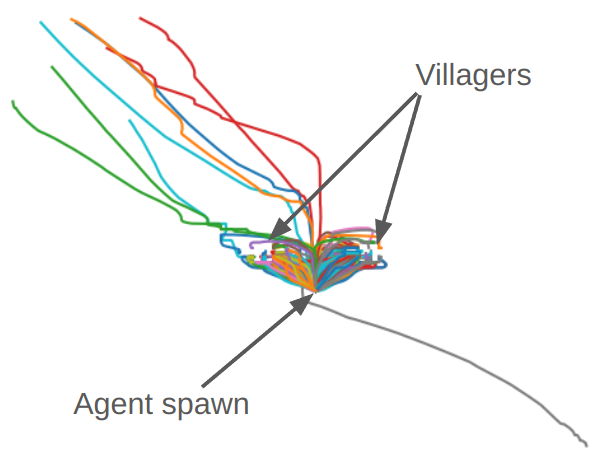}
    \caption{Top-down view of trajectories when VPT is presented with a choice between two villagers standing under tree leaves, one on the left, one on the right. The identical scenarios produce different trajectories due to stochastic actions, including one trajectory where the agent turns around and goes in the other direction.}
    \label{fig:top-down-coords-two-villager-trees}
  \end{minipage}
\end{figure*}

In addition to visualizing attention weights (\autoref{fig:attention-over-time}), we can also gain insights by visualizing attention outputs.
For instance, attention head 2.2 again shows a distinct pattern when the camera is moved up (see \autoref{fig:activation-viewer-2.2}).
We discovered some of the patterns in the attention weight plots after initially noticing them in the attention output visualizations.

Below we list some of the patterns we found in both the attention weight and the attention output visualizations:

\begin{itemize}
    \item Attention head 0.8---looks at the previous frame every time the inventory menu is open (\autoref{fig:all-heads-layer-0}, \autoref{fig:activation-viewer-1.2}).
    \item Attention head 0.9---looks at the previous frame all the time, except for each time when the inventory menu state is changing, when it looks at the current frame (\autoref{fig:all-heads-layer-0}).
    \item Attention head 1.2---mostly looks at every 4th frame (\autoref{fig:all-heads-layer-1}, \autoref{fig:activation-viewer-1.2}).
    \item Attention head 2.2---looks at the current frame when the agent is looking up (\autoref{fig:all-heads-layer-0}).
    \item Attention head 3.3---shows a distinct pattern when the agent decides that it needs to open the crafting table menu (\autoref{fig:activation-viewer-3.3}).
\end{itemize}

While visualizations can suggest what certain attention heads are doing, one of the main goals is to be able to control the agent's behavior in a predictable manner, which we explore in the next section.

%% file: sections/50_intervention.tex
 \section{Interventions and ablations}

After visualizing the attention layers and observing the agent, we wanted to better understand the behavior and possibly change it by performing various interventions and ablations.
We start with simple behavioral interventions in which we put the agent in different situations and observe how it acts.
We then show how such observations are not very informative unless hundreds or even thousands of episodes are run.
After this, we discuss the metrics we used for further experimentation: visual manipulations and ablations of the attention parts of the agent network.

\subsection{Behavioral interventions}
We have performed various behavioral experiments to get better intuitions about how the agent makes decisions.
One should view these in the spirit of Eugene Linden \citep{Linden1999, Linden2003}---whose books contain many stories of surprising animal behavior---as anecdotes and not as proof of some mechanism of behavior.

First, we wanted to see if VPT would be tricked in a deceptive situation.
We gave the agent an iron pickaxe at the start of an episode and placed it in front of a tree and a block of diamond ore (see \autoref{fig:dmd-ore-vs-tree} in the appendix).
Mining the diamond ore gives a 20 times higher reward, but chopping the tree is what the agent usually does at the start of an episode.
In the first rollout, it chose the tree, while in the second it chose the diamond ore.
What does that say about how the agent makes decisions?

We then performed an experiment to better understand how the agent decides that it is time to start making a crafting table.
Does it do it after chopping trees for a while?
Or maybe when seeing the top log of a tree from below?
To find out, we gave it enough logs to make a crafting table at the start of an episode.
The agent ran around for a bit, opened its inventory, moved the mouse to the logs, picked them up... then dropped them and moved on to find some trees to chop.
That does help getting back into familiar territory, but it might not be the smartest behavior.
This happened on the very first try before we started recording.
We then tried to reproduce it for a video and it took over 500 episodes before it happened again (see \href{https://youtu.be/e5qWNVEtuDA}{Video05}).
In most of the remaining episodes, it crafted the logs and proceeded to solve the task.

Finally, we discovered the scenario in which VPT kills a villager by using the knowledge that vision models often use spurious features to recognize objects \citep{izmailov2022feature}.
This also happens in biological systems, such as the Australian jewel beetle males mistaking brown beer bottles for females \citep{gwynne1983beetles}.
First, we tried finding a natural scenario, such as some villagers standing under tree leaves in a village, where VPT would attack them.
This happened sometimes, but the villagers would quickly run away and the agent would lose interest in attacking them.
We thus forced a villager to stand in one spot by placing four invisible barrier blocks around it and some tree leaves and logs above.
This also meant the agent could not move closer to the villager than the barrier blocks allowed.
In this scenario, the agent punches the villager to death, which takes around 20 punches, in roughly 30\% of episodes.
This last example, although contrived, illustrates the necessity of a better understanding of AI agents, if we are to trust them in real-world situations.

\subsection{Stochastic actions and metrics}

In the previous section, we saw how difficult it can be to interpret agent behavior.
Especially when the actions are stochastic.
The agent network outputs action probabilities, which are then sampled with a default temperature parameter of 1.
When the agent sees a tree right in front of it, it is very certain of what it should do---stand still and attack.
The attack probability in this case is often above 99.99\%.
However, if we present it with a less certain Y-maze type scenario of two villagers under leaves an equal distance away on both sides, it will be less certain.
The camera action probabilities in such a case could be 40\% left, 40\% right, and 20\% no change, for example.
This means that each time we roll out the agent in an identical situation, we can have a different outcome.

To illustrate (see \autoref{fig:top-down-coords-two-villager-trees}), out of roughly 100 rollouts with the two villagers, roughly half the time the agent went for the left one, and the other half it went for the right one.
However, there were several cases, where the agent ignored both villagers and ran past them towards some trees further away on the left side.
Finally, there was a single trajectory where the agent turned around and went on its merry way in the opposite direction.
This makes it hard to perform detailed ablation analysis because each ablation would require hundreds or even thousands of rollouts to see if there was any behavior change (A single rollout of VPT takes roughly 15 CPU minutes).
Even then, it would not be enough to look at summary statistics to calculate the change in behavior---it might be the case that the change happens only in rare circumstances.
Because of this, we use probability differences and log probability differences of actions as the main metrics for our ablation experiments.
Both metrics have their strengths and weaknesses, which are described well by \citet{heimersheim2024activation}.

\subsection{Visual manipulation}

Many successes of interpretability work in large language models have been achieved by carefully crafting the prompts and inspecting the resulting attention patterns.
For example, induction heads have been identified by giving a repeated random sequence of letters to GPT-2 \citep{olsson2022incontext}.
Inspired by this discovery, we gave the agent modified sequences of observations and inspected the resulting attention patterns.

For example, when we repeat the first frame of the episode 1,000 times, the attention head 1.2 still shows the pattern of paying attention to every fourth frame (see \autoref{fig:head-1.2-repeated-frame-0} in the appendix).
There was no frame-skip or frame-stack of four during training, which is often used in other RL agents.
This means that the phenomenon is not purely a reflection of the observations but reveals something about the agent itself.

\begin{figure*}[htbp]
  \centering
  \includegraphics[width=\textwidth]{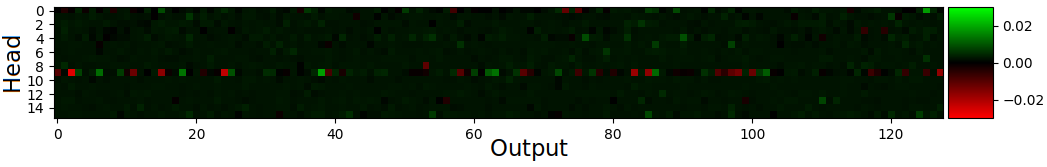}
  \caption{
  The change in attack probability after ablating one-by-one every output in each attention head in layer 0 for frame 20 in a regular episode.
  The highest impact ablations are in attention head 0.9 (bright horizontal line).
  See \href{https://youtu.be/ju-s301cHzI}{Video07}.}
  \label{fig:single-output-ablation-heatmap-frame-20}
\end{figure*}

Another experiment was inspired by the infamous 25th frame effect, the discredited belief that subliminal messages could be embedded into a film by inserting them into every 25th frame, influencing viewers' thoughts and behaviors without their conscious awareness.
We showed the inventory screen for a single frame at the 50th time step of an episode and left the remaining observations unchanged.
This caused attention head 0.1 (and some others too, but in a less pronounced way) to start focusing mainly on this frame as long as it could (see \autoref{fig:head-0.1-50th-frame-replaced-by-inventory} in the appendix).
Interestingly, the resulting action probabilities were changed beyond the 128-frame window of its memory.
This is because the hidden state of the transformer could keep some information from those distant frames.
However, these long-range changes were minor, on the order of 0.1\%.

For a scenario closer to reality, we tried imagining how we could intervene in the observations of an embodied agent in the real world instead of Minecraft.
One idea was to shine a laser pointer at the agent's camera.
Thus, we replaced 128 frames with pure red color starting at frame 150.
This in effect wipes the memory of the agent.
The attention patterns did not reveal much; however, the action probabilities did.
Attack probability returned to nearly 100\% within a few frames, but the less certain camera direction probabilities took much longer to recover.
This could be simply because of the particular point in time we chose for this intervention---the agent was looking at the tree and did not need its memory to figure out what it was doing.
However, this could also mean that uncertain actions are easier to influence with such interventions.
We demonstrate more evidence for this hypothesis in the next subsection.

\subsection{Attention ablations}
\label{subsection:attention-ablations}

One simple but informative experiment is to run the agent without access to its memory.
We implement this by resetting the transformer's hidden state to its initial condition at every time step and letting it observe only the current frame.
It barely manages to run towards a tree and painstakingly chops a single log (see \href{https://youtu.be/0-bxLngYO1Y}{Video06}).
It fails to craft altogether.
This shows the importance of memory for VPT even when a single frame is enough to infer the next action.

In prior work on interpreting VPT, \citet{joseph2023mining} mean-ablate each attention head one-by-one.
They find that ablating attention head 0.9 results in a significant logit difference for attack versus non-attack actions.
Despite the logit difference, ablating the identified head does not impact the agent's performance in terms of task completion.

To see if we can find a more granular way of influencing agent actions we ran the following experiment.
We recorded a 15-second (300 frames) episode where the agent runs towards the villager-tree, punches the villager for about 3 seconds, and then looks up to chop the logs instead.
We record the action probabilities for each frame.
They stay constant if we show the same sequence of frames to the agent again, making the experiment fully reproducible.
We start with the first frame.
We take the first of the 128 outputs of attention head 0 in layer 0 and zero-ablate it, i.e. set the activation value to zero.
We did not notice large differences between zero and mean ablations, so we chose zero ablations for simplicity.
We record the resulting modified action probabilities, set the output back to the previous value, and move to the second output of the same attention head.
We do this for each output in each attention head in each layer for a total of 128*16*4=8,192 times.
We do this for each frame.
This lets us know which outputs have a higher influence on the resulting agent actions.
We visualize the heatmap for the 20th frame for the 16 attention heads in layer 0 in \autoref{fig:single-output-ablation-heatmap-frame-20}.
We can see that the most impactful outputs in layer 0 belong to head 0.9, which coincides with the findings by \citet{joseph2023mining}.

\begin{figure*}[htbp]
\centering
\begin{minipage}{.48\textwidth}
  \centering
  \includegraphics[width=.99\linewidth]{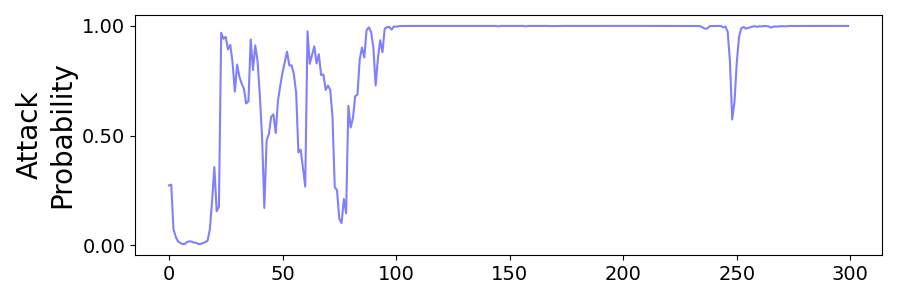}
  
  \vspace{0.5em}
  \includegraphics[width=.99\linewidth]{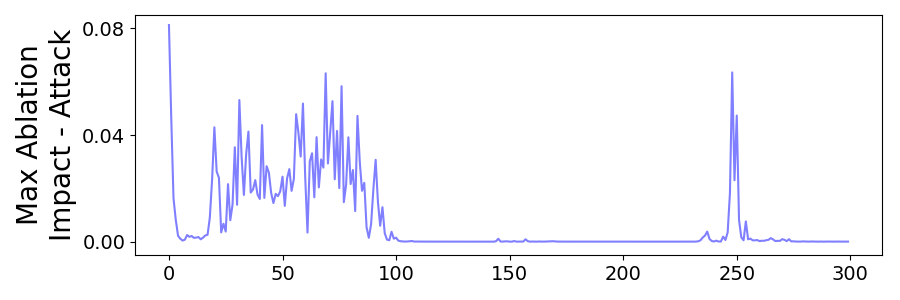}
  
  \vspace{0.5em}
  \hspace{2.6em}
  \includegraphics[width=.78\linewidth]{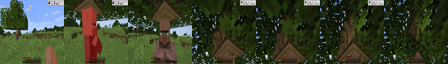}
\end{minipage}
\begin{minipage}{.48\textwidth}
  \centering
  \includegraphics[width=.99\linewidth]{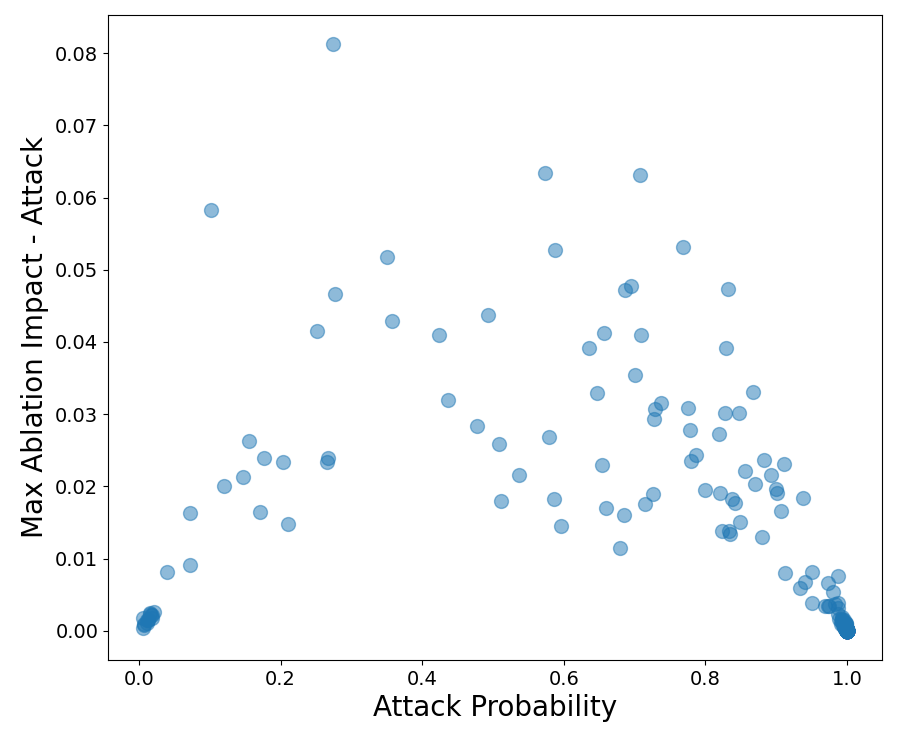}
\end{minipage}
\caption{
(Top left) Attack probability over a 300-frame episode where VPT attacks a villager-tree for about 100 frames, after which it looks up to chop tree logs.
(Middle left) Maximum single output ablation impact over the episode---zero impact when attack probability is certain (either 0\% or 100\%).
(Bottom left) a visualization of the episode.
(Right) Same information as a scatter plot, showing how uncertain attack probabilities coincide with higher ablation impact.
See \href{https://youtu.be/ju-s301cHzI}{Video07}.}
\label{fig:attack-probability-vs-max-ablation-impact-and-attack-scatter}
\end{figure*}

We then compare the maximum ablation impact for each frame to the unablated action probabilities (see \autoref{fig:attack-probability-vs-max-ablation-impact-and-attack-scatter}).
The only times when we can produce a meaningful impact using this method is when the agent is uncertain about its actions.
When the probability of attack is near 0\% or 100\%, such ablations do nothing.
This is not simply the effect of probability scaling exponentially with the logits---we observe an almost identical phenomenon if we replace the probability differences with log probability differences (see \autoref{fig:probabilities-vs-log-probabilities} in the appendix).
We chose to display raw probabilities for clarity, but there are also good reasons to use other metrics instead \citep{heimersheim2024activation}.

The high certainty of the attack action also suggests a potential fix to the murderous agent problem.
Namely, do not let the agent execute an attack action unless it is highly certain of it.
We performed a preliminary experiment, where we set the threshold to 99\%.
The agent stopped randomly attacking the air while running around, it also did not attack the villager-tree.
It carefully, after looking at it for a moment, chopped a tree, made a wooden pickaxe, and even destroyed and picked up the crafting table.
However, it got stuck trying to dig down through the grass afterward (See \href{https://youtu.be/U8NYiudY5n8}{Video08}).
The fix prevented the worst misbehavior but also made the agent incapable of solving its task.

%% file: sections/60_discussion_and_limitations.tex
\section{Discussion and limitations}

There is an obvious limitation to our work.
We are like a behavioral biology lab with a single rat.
What claims can we make about any agent other than VPT?
In fact, would any of our analyses transfer beyond the single checkpoint we analyzed?
We briefly experimented with STEVE-1 \citep{lifshitz2023steve1} and easily succeeded in making it chop village houses, so this type of behavior might be common (see \autoref{fig:steve-1-housechop}).
However, we know from prior work \citep{sellam2021multiberts, chughtai2023toy, damour2022underspecification} that factors such as the random seed used for weight initialization can have a large influence on how models learn to solve specific tasks.
This is the case even when fine-tuning models instead of full retraining.
It shows up as similar in-distribution performance but differing behavior in out-of-distribution situations.
Situations such as the villager standing under some leaves.
It is likely that if VPT was retrained many times with only the random seed used for weight initialization differing between runs, the resulting agents could exhibit different behavior when presented with the villager-tree.

Would that make our analysis useless?
The goal of this work was to show that it is possible to use existing interpretability techniques to discover insights into the behavior of a complex vision-based agent with a transformer component.
It also provides ideas for how other similar agents might make decisions---ideas that can be tested.

However, it does raise an additional consideration---if the internal mechanisms of an agent can change unpredictably during training, our method of manually applying these techniques would not scale well.
We would then need interpretability techniques that can be easily and automatically applied after each significant change of a deployed model.

Another limitation of our work is that we considered only single attention head output ablations.
Much success in mechanistic interpretability of LLMs happened through discovering circuits by ablating the connections between attention heads 
instead \citep{wang2022interpretability}.
We believe this points to a fruitful future research direction.

%% file: sections/90_conclusion.tex
\section{Conclusion}

This study advances our understanding of decision-making in VPT, a large vision-based reinforcement learning agent.
By applying interpretability techniques, we have provided clues on how the agent manages complex tasks with limited memory, focusing on recent and key-frames to maintain task coherence.
Our findings include a new example of goal misgeneralization in the wild, where the agent mistakenly identified a villager as a tree trunk and punched it to death, pointing to the urgent need for better interpretability and error correction in AI systems.
This research underscores the critical role of interpretability in ensuring the safety and transparency of AI models.

%% file: sections/100_appendix.tex
\section{Saliency maps experiments}
Although saliency maps are known to have many issues, they do serve as an additional test to validate some of our previous hypotheses.
We implement Gradient \citep{simonyan2014deep} and SmoothGrad \citep{smilkov2017smoothgrad} saliency maps.
When applied to VPT, both Gradient, and SmoothGrad pass the sanity checks specified in \citet{adebayo2018sanity}: the model parameter randomization test and the data randomization test.

\begin{figure}[htbp]
  \centering
  \begin{minipage}[t]{0.45\textwidth}
    \includegraphics[width=\textwidth]{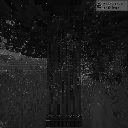}
    \caption{
    Gradient saliency maps on the tree in between VPT's punches.
    Colors are inverted to make it easier to see the contrast between the saliency maps and the gameplay.
    See \href{https://youtu.be/cCRGOTRZQ8U}{Video09}.}
    \label{fig:saliency1}
  \end{minipage}
  \hfill
  \begin{minipage}[t]{0.45\textwidth}
    \includegraphics[width=\textwidth]{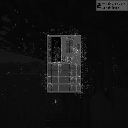}
    \caption{
    VPT opens up its inventory and the gradient saliency maps indicate that it is looking at the first slot in the hotbar, off-hand slot, and the crafting area.
    However, in this rollout, VPT did not have anything in the inventory.
    VPT normally checks its inventory after attacking a tree, but in this case, it attacked the villager-tree and then checked its inventory which was empty.}
    \label{fig:saliency2}
  \end{minipage}
\end{figure}

The saliency maps of VPT show that when the agent is chopping logs, it focuses on the tree (see \autoref{fig:saliency1}).
From afar, VPT appears to look more at the foliage of the tree than at the tree trunk itself.
When VPT is confronted with the tree vs villager-tree dilemma, slightly more focus is paid to the real tree than to the villager-tree.
The saliency map of the inventory screen can be seen in \autoref{fig:saliency2}.

\newpage
\section{CNN Feature visualization}

\subsection{Filter Visualization by Optimization}
\begin{figure}[htbp]
  \centering
  \includegraphics[width=0.6\columnwidth]{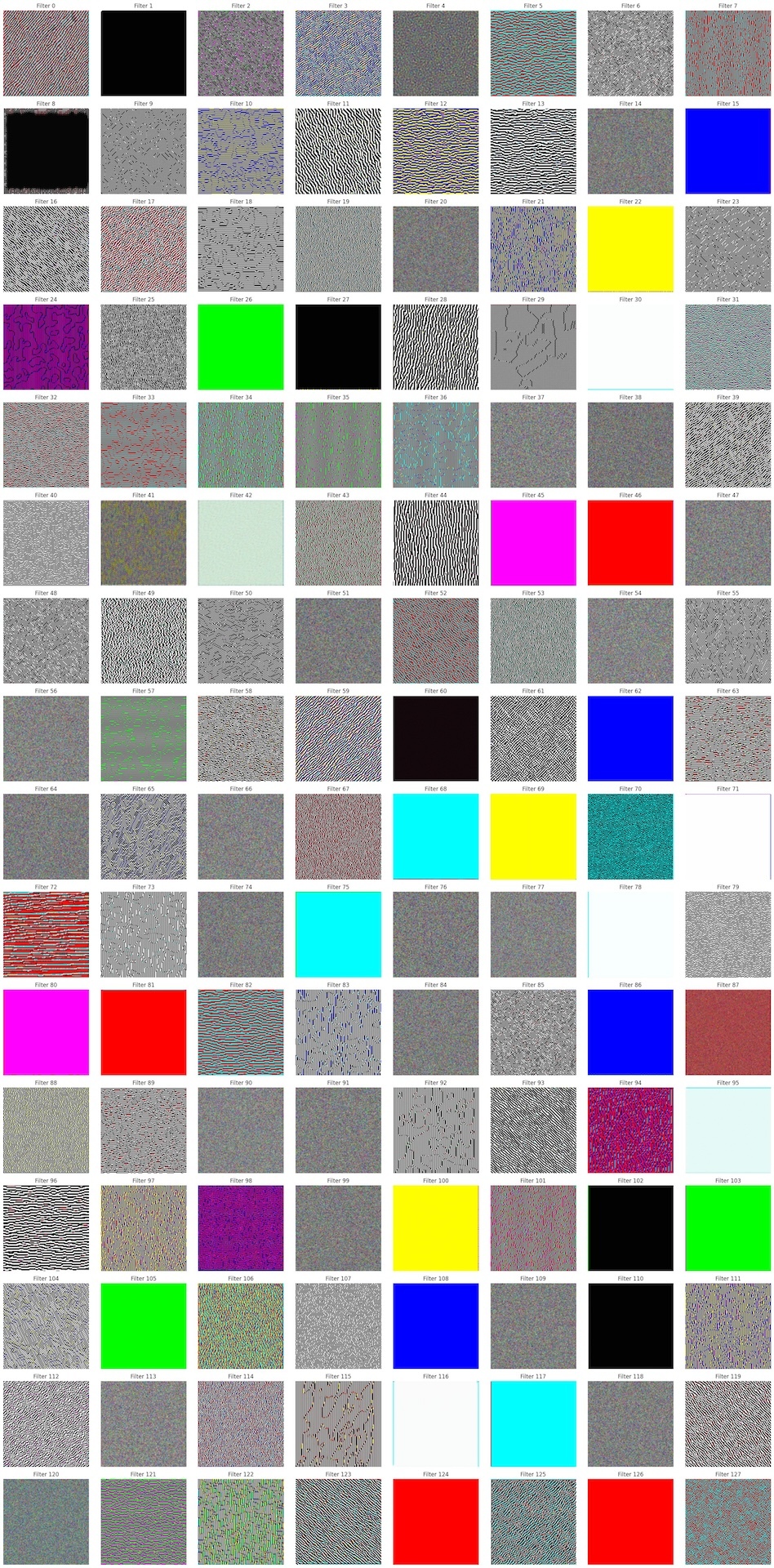}
  \caption{Filter visualization by optimization for the 5th CNN layer of VPT.}
  \label{filter_vis6}
\end{figure}

\begin{figure}[htbp]
  \centering
  \includegraphics[width=0.6\columnwidth]{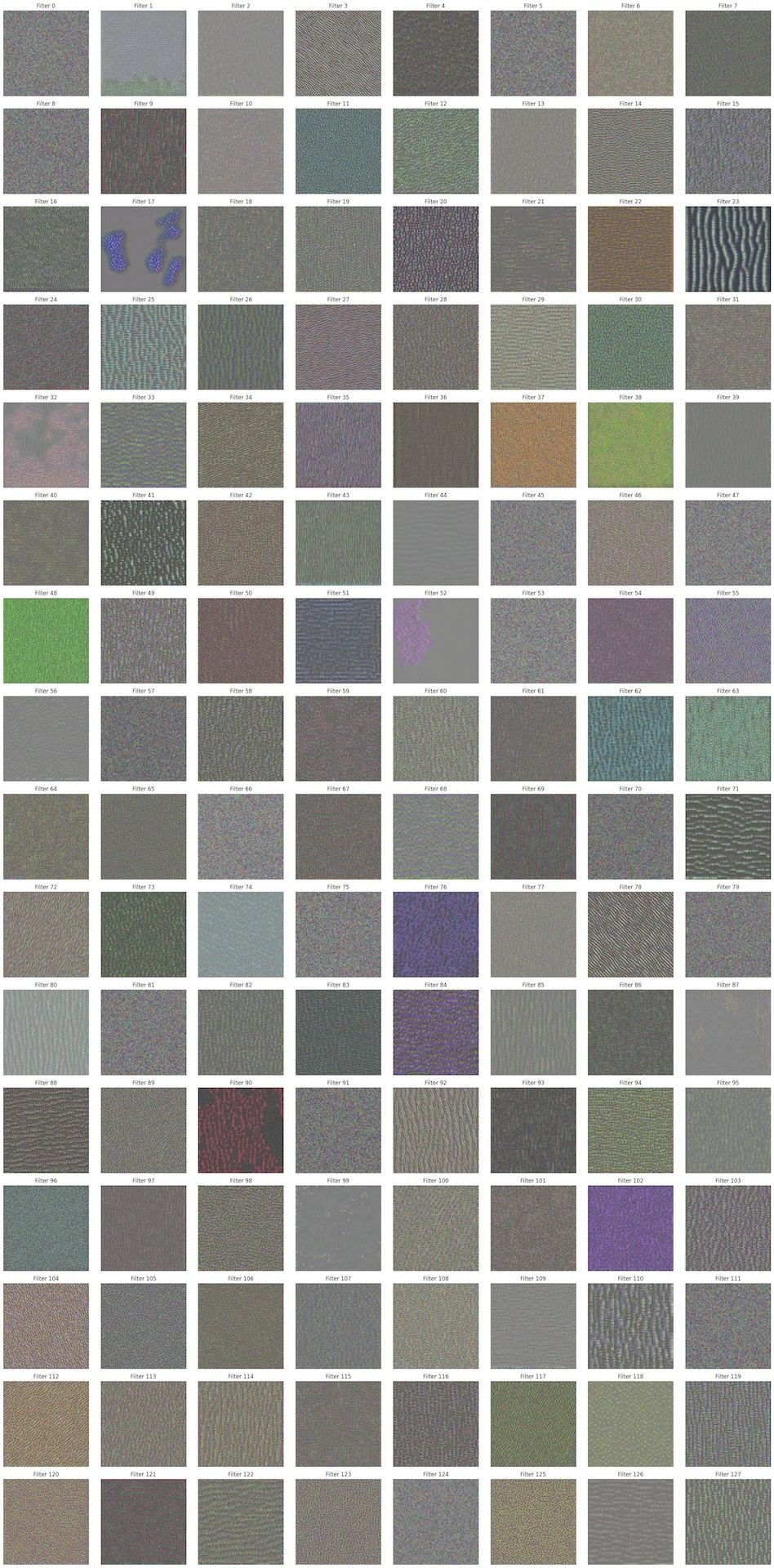}
  \caption{Filter visualization by optimization for the 1st CNN layer of VPT.}
  \label{filter_vis_0}
\end{figure}

Our study leverages filter visualization by optimization to elucidate the specific patterns and features that various layers of the VPT model detect.
The goal of this optimization is to generate an image that maximizes the activation of a specific kernel map.
The loss function is the output of that kernel map.
In \autoref{filter_vis6}, we show the visualizations of the fifth CNN layer's filters.
These visualizations reveal that filters at this depth in the network capture more complex textures and patterns compared to initial layers (see \autoref{filter_vis_0}).

\autoref{filter_vis_0} provides filter visualizations for the first CNN layer.
These filters detect basic features such as edges, simple textures, and colors.
The simplicity of these features aligns with the layer's role in capturing foundational visual elements, which are progressively combined and abstracted in the deeper layers to form more complex representations.

\subsection{Receptive Field Attention}
\begin{figure}[htbp]
  \centering
  \includegraphics[width=0.5\columnwidth]{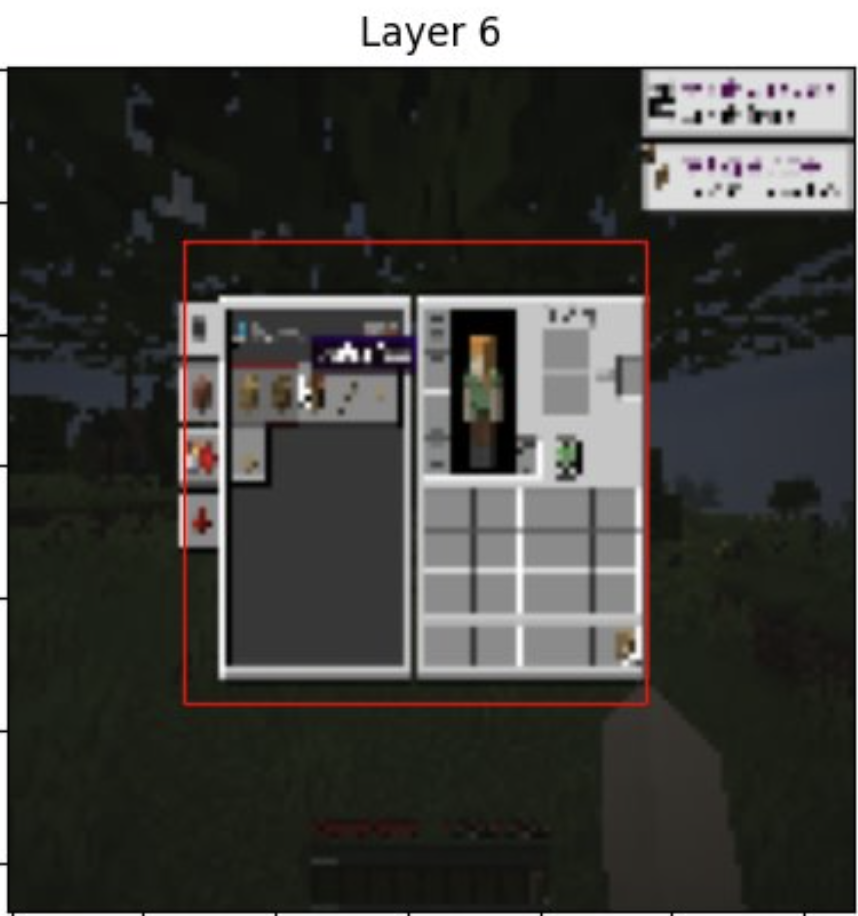}
  \caption{Receptive field attention of layer 6 mapped into an input image.}
  \label{receptive_rect}
\end{figure}

\autoref{receptive_rect} shows the receptive field attention of layer 6 mapped onto an input image.
This mapping shows how the activations from this layer correspond to specific regions in the input image, highlighting areas of interest that the model focuses on during gameplay.

The receptive field \(R\) of a convolutional layer refers to the region in the input image that affects a particular feature in the output.
It can be calculated using the kernel size \(K\), the stride \(S\), and the padding \(P\). The basic formula for a single layer is straightforward:
\[
R = K
\]

For deeper layers, the calculation becomes recursive. Let’s denote:
\begin{itemize}
    \item \(R_L\) as the receptive field of layer \(L\)
    \item \(K_L\) as the kernel size of layer \(L\)
    \item \(S_L\) as the stride of layer \(L\)
    \item \(P_L\) as the padding of layer \(L\)
\end{itemize}

The receptive field of a layer \(L\) depends on the receptive field of the previous layer \(R_{L-1}\) and can be calculated as:
\[
R_L = R_{L-1} + (K_L - 1) \times \prod_{i=1}^{L} S_i
\]

Where \(\prod_{i=1}^{L} S_i\) is the product of all the strides from layer 1 to layer \(L\).

This recursive method allows us to determine the receptive field of any layer in a deep convolutional network by considering the kernel size, stride, and padding of all preceding layers.
The mapping of activations to specific regions in the input image (like gameplay frames) helps in understanding which parts of the input the model focuses on.
For instance, the activations in a higher layer of the network will correspond to larger regions of the input image, indicating broader, more abstract features such as objects or actions within the game.
In \autoref{receptive_heatmap}, we present heatmaps of the receptive fields associated with the top 500 activations of the sixth layer.

\subsection{Kernel Visualization}
\begin{figure}[htbp]
  \centering
  \begin{minipage}[t]{0.48\textwidth}
    \vspace{0pt}
    \includegraphics[width=\textwidth]{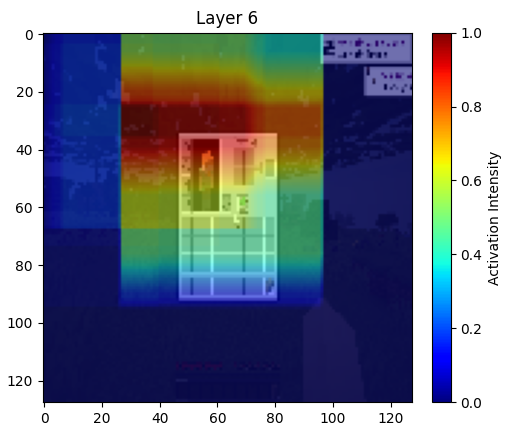}
    \caption{Receptive fields associated to top 500 activations of 6th layer.
    In other words, the receptive field attention of layer 6 mapped into an input image.}
    \label{receptive_heatmap}
  \end{minipage}\hfill
  \begin{minipage}[t]{0.48\textwidth}
    \vspace{0pt}
    \includegraphics[width=\textwidth]{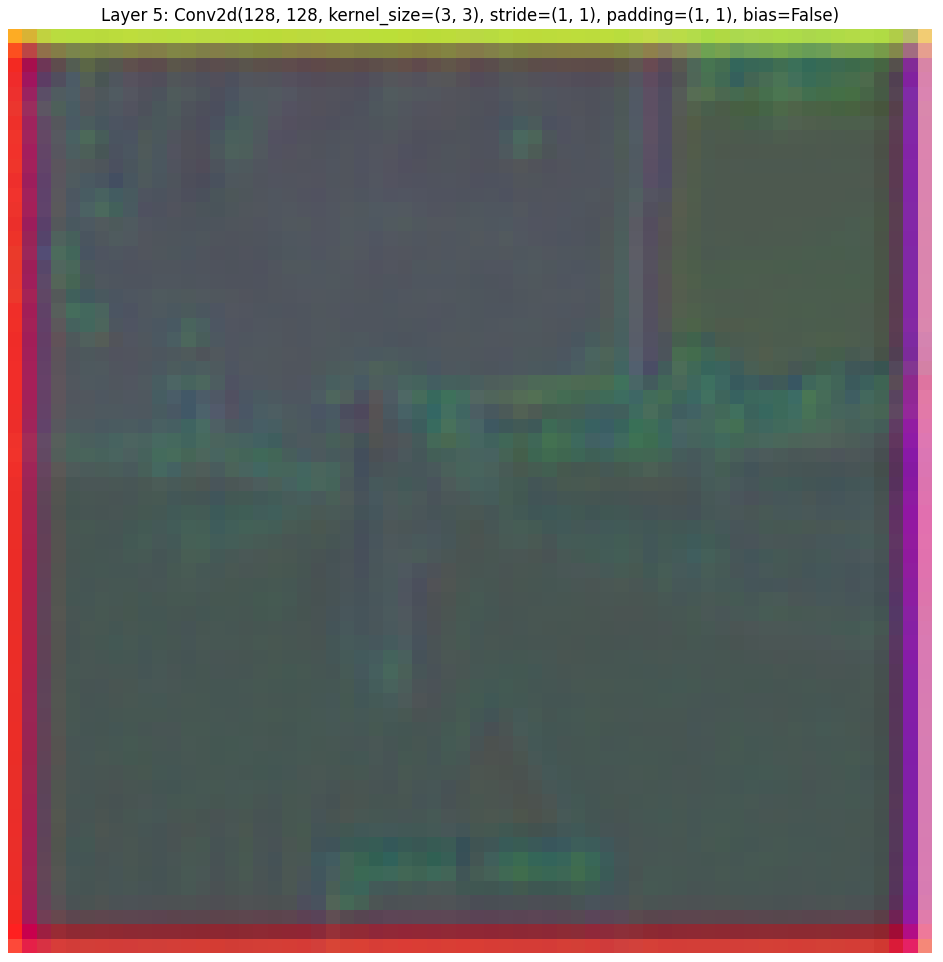}
    \caption{How kernels visualize the image for VPT in the 5th layer, this is the villager-tree input image and lack of boundary in the image suggests that VPT didn't highlight the difference between the villager and tree.}
    \label{kernel_vis5}
  \end{minipage}
\end{figure}

The kernel visualization for the fifth layer, shown in \autoref{kernel_vis5}, demonstrates how the model interprets an image containing a villager and a tree.
For this figure, we combined the outputs of multiple kernels and applied PCA to reduce the dimensionality to three principal components.
This allowed us to visualize complex, high-dimensional features in an RGB format, highlighting how different components contribute to the learned representations.
This technique revealed the intricate patterns learned by the network and how different kernels collaborate to encode features.
Interestingly, the visualization indicates that the model does not distinctly separate the villager from the tree, suggesting a limitation in the model's ability to discern boundaries between overlapping objects.

\subsection{Filters Overlay}
\begin{figure}[htbp]
  \centering
  \includegraphics[width=0.9\columnwidth]{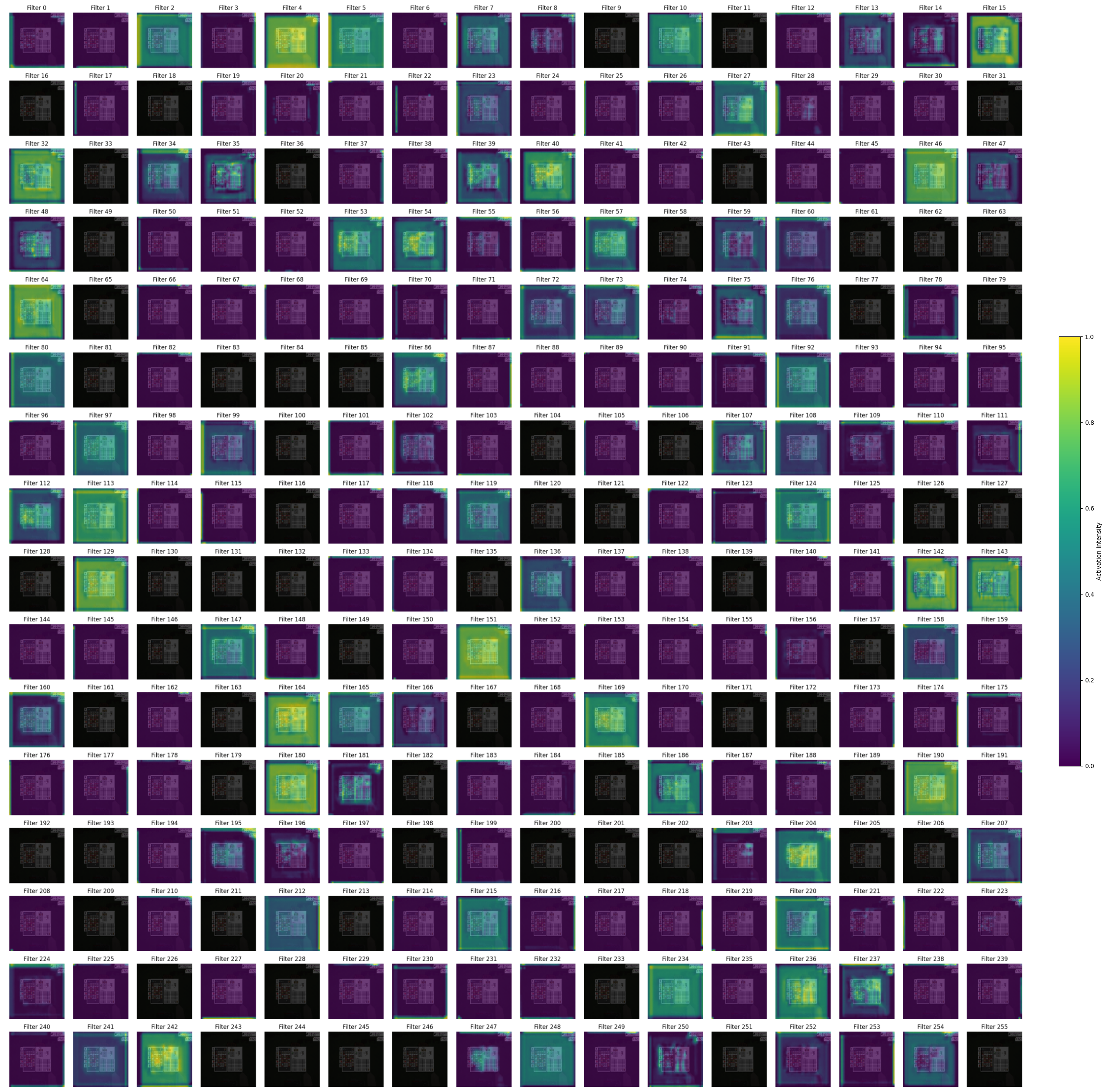}
  \caption{Filters overlay.}
  \label{filter_overlay}
\end{figure}

\autoref{filter_overlay} showcases the overlay of filter activations, providing a comprehensive view of how different filters respond to the same input image. Here's how we generate these filter overlays
\begin{itemize}
    \item Pass an image through the network: the input image is fed through the CNN, and the activations of a specific layer are recorded.
    \item Extract filter activations: For the chosen layer, the activations of each filter are extracted. These activations are essentially the feature maps generated by each filter when applied to the input image.

    \item Overlay the activations on the input image: Each filter's activation map is overlaid on top of the input image. This can be done by upscaling the activation maps to the original image size and blending them with the image using different visualization techniques like heatmaps or transparency overlays (we used transparency overlays).
\end{itemize}
This overlay helps understand the cumulative effect of multiple filters working together, contributing to the overall feature extraction and decision-making process in the VPT model.

\subsection{Layer-Specific Activations}
\begin{figure}[htbp]
  \centering
  \includegraphics[width=0.9\columnwidth]{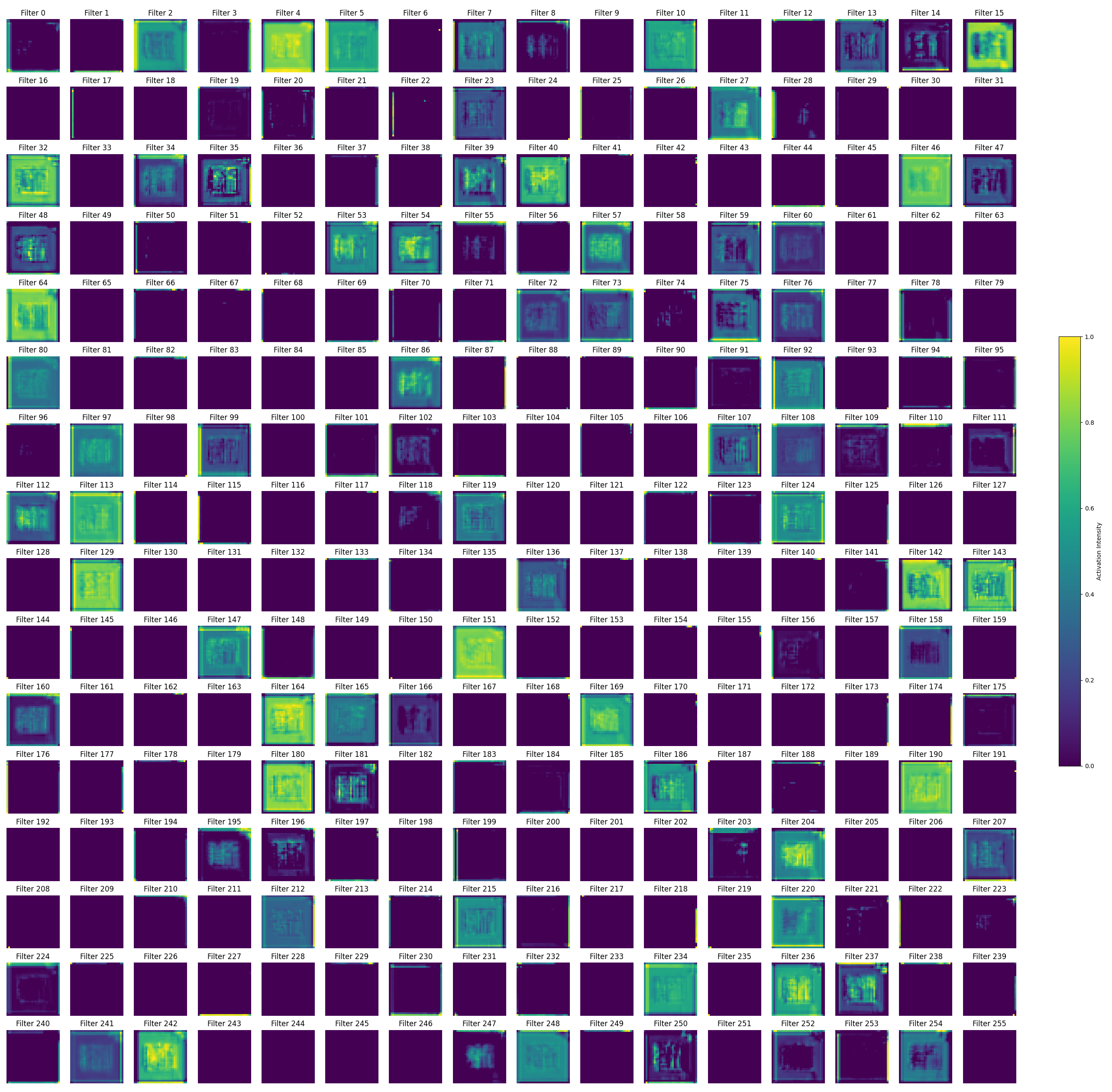}
  \caption{Kernel visualization for an inventory image of the 6th layer.}
  \label{kernel_vis}
\end{figure}

The kernel visualization of all 256 kernels for an inventory image in the sixth layer, depicted in \autoref{kernel_vis}, emphasizes how the model processes and prioritizes different parts of the inventory interface.
The activations suggest that the model pays particular attention to areas with high informational content, such as item slots and inventory shapes (filters 15, 35, 39, 142, 143).

\clearpage
\newpage
\section{Representation engineering}

\citet{zou2023representation} introduced the idea of running forward passes through a model with a specific concept in mind, then saving the activations of some layer in the model for both the positive and negative versions of that idea.
For example, if we want VPT to be more likely to attack, we can find a way to add the activations of the positive concept of a tree and subtract the activations of the negative concept of an empty field.
Then, modify the model so that during inference, it adds the positive version of a concept and subtracts the negative version of that concept from the same layer.
The model's outputs tend to be steered in the direction of your concept.
This technique has been used to steer the behavior of GPT-2 \citep{zou2023representation} and to control a maze-solving RL agent \citet{mini2023understanding}.
We apply this to VPT with some limited success.

\begin{figure}[htbp]
    \centering
    \begin{subfigure}[b]{0.45\textwidth}
        \centering
        \includegraphics[width=\textwidth]{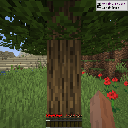}
        \caption{Tree Image}
        \label{fig:tree}
    \end{subfigure}
    \hfill
    \begin{subfigure}[b]{0.45\textwidth}
        \centering
        \includegraphics[width=\textwidth]{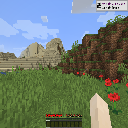}
        \caption{Blank Image}
        \label{fig:blank}
    \end{subfigure}
    \caption{These are the images used when attempting to steer VPT towards being more prone to attacking using techniques of representation engineering.
    We add 3x the first MLP layer's activations of the tree image minus 3x the first MLP layer's activation when in the field without trees.}
    \label{fig:representation_3x_side_by_side}
\end{figure}

We modified the VPT model by targeting the first MLP layer in the first transformer block after the CNN. We added 3x the first MLP layer's activations when VPT was in front of a tree, and subtracted 3x the first MLP layer's activations when VPT was in a field with no trees, i.e. +3*(tree image - field image) (see \autoref{fig:representation_3x_side_by_side}).
The modified VPT kept punching at thin air even if there were no trees around (see \href{https://youtu.be/i4RbOqFDlKc}{Video10}).
However, this change caused VPT to lose its ability to achieve goals it was previously capable of accomplishing.

Conversely, when we changed the sign of the activation engineering, i.e. -3*(tree image - field image), the modified VPT did not punch a tree that was right in front of it (see \href{https://youtu.be/8NcUdqmCY4k}{Video11}).

As found in \citet{turner2024activation}, the scalar value with which to multiply a layer's activations before adding/subtracting them from the model's forward pass is mostly guess and check.
We tried scalar values between 1 and 10, and of these, 3 worked the best.
We also find that this technique breaks if the magnitudes of the scalars multiplying the positive/negative activation vectors are not equal. 

A more fine-grained approach such as \citet{templeton2024scaling} could potentially be useful in attempting to not "break" the model when trying to steer it towards certain kinds of actions.

\clearpage
\newpage
\section{Other}

Below we show more figures that were not included in the main body due to space constraints.

\begin{figure}[ht]
  \centering
  \includegraphics[width=\linewidth]{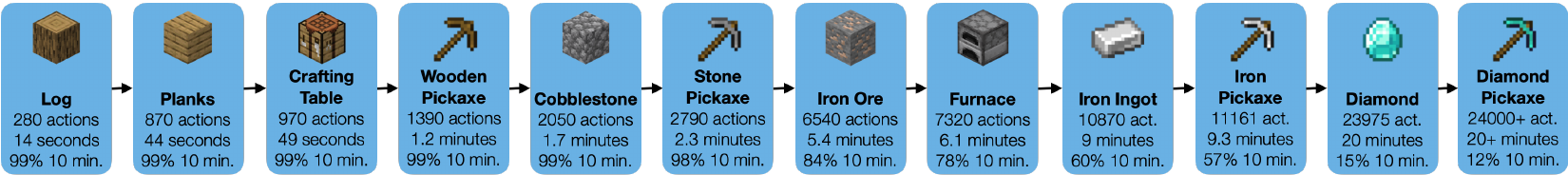}
  \caption{Sequence of subtasks to get a diamond pickaxe in Minecraft. Figure from \citet{baker2022video}.}
  \label{fig:curriculum}
\end{figure}

\begin{figure}[htbp]
  \centering
  \includegraphics[width=\linewidth]{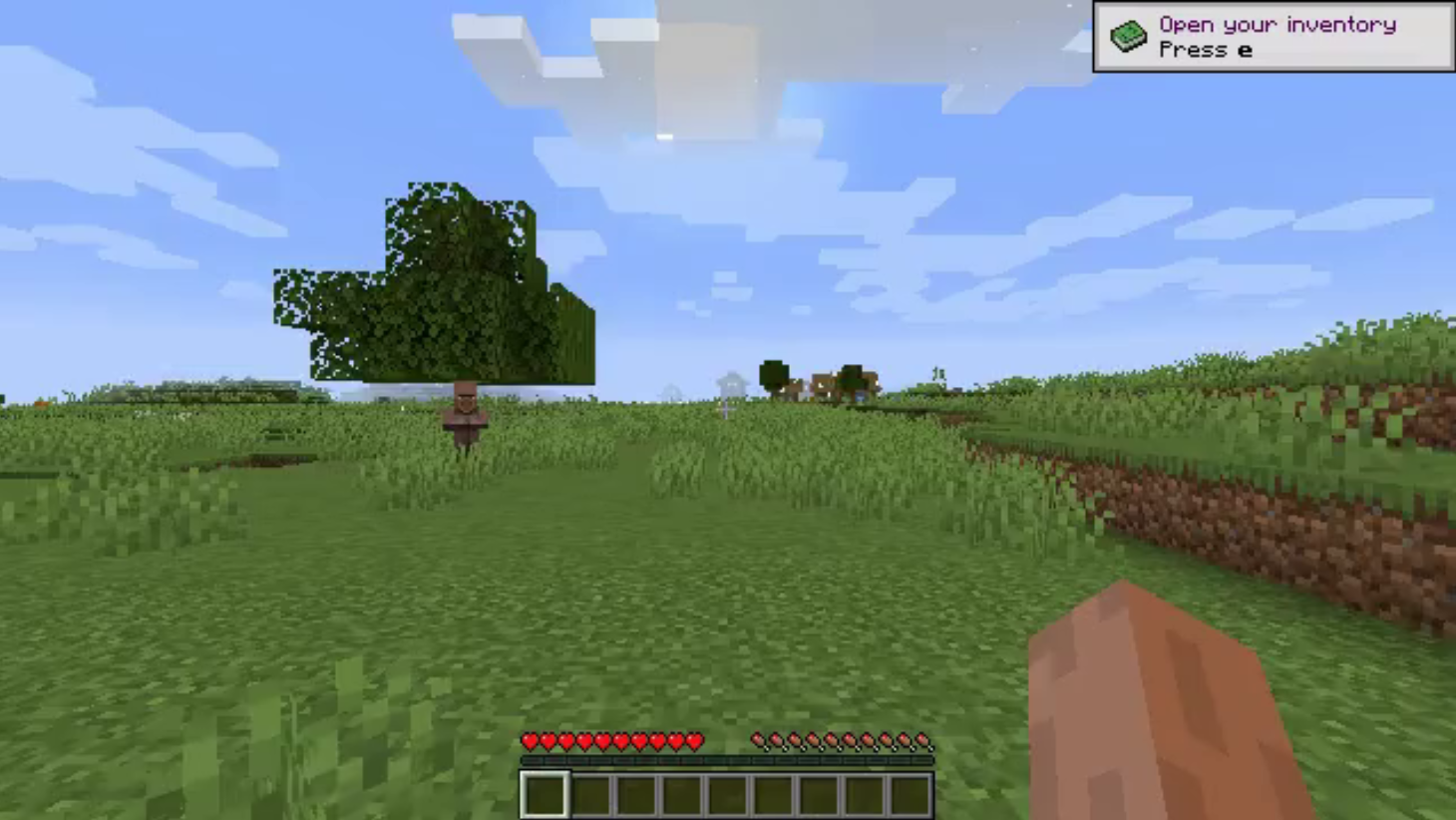}
  \caption{
  A "villager-tree".
  We replaced the trunk of an oak tree with a villager and made it stand in one place by placing four invisible barrier blocks around it.
  The agent goes for the villager-tree most of the time, punches it multiple times, and often kills the villager, which takes around 20 punches.
  See \href{https://youtu.be/VVkWWgwKf0M}{Video12}.}
  \label{fig:villager-tree-example}
\end{figure}

\begin{figure}[htbp]
  \centering
  \includegraphics[width=\linewidth]{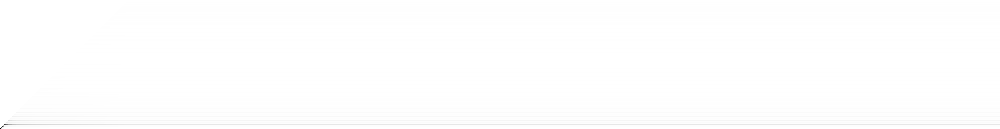}
  \caption{
  Attention weights of attention head 1.2.
  The first frame of the episode is repeated 1000 times.
  The pattern of paying attention to every fourth frame is still visible.}
  \label{fig:head-1.2-repeated-frame-0}
\end{figure}

\begin{figure}[htbp]
  \centering
  \includegraphics[width=\linewidth]{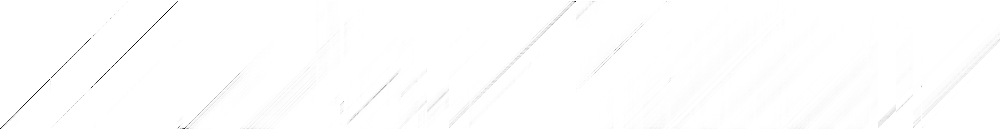}
  \caption{
  Attention weights of attention head 0.1.
  The 50th frame is replaced by the 316th frame, which shows open inventory.
  This attention head pays full attention to the replaced frame for most of the maximum 128-frame duration (2nd bright diagonal line on the left side).}
  \label{fig:head-0.1-50th-frame-replaced-by-inventory}
\end{figure}

\begin{figure}[htbp]
  \centering
  \includegraphics[width=\linewidth]{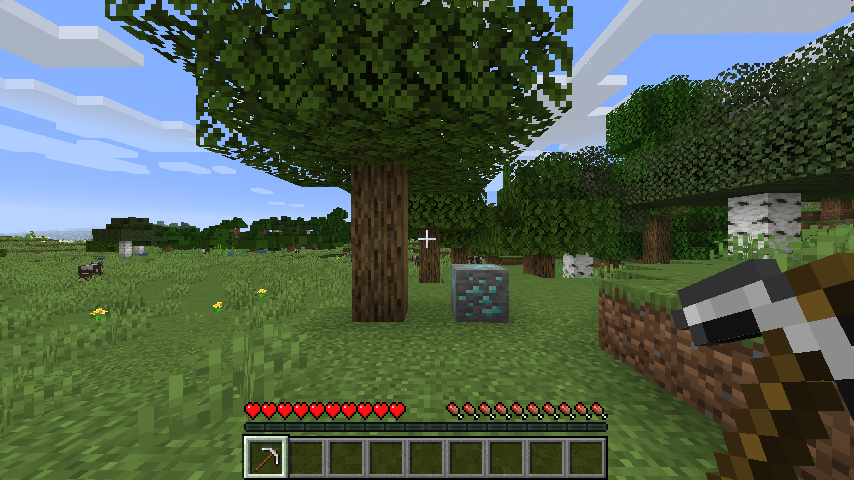}
  \caption{
  The agent is given an iron pickaxe and has to choose between mining diamond ore and chopping a tree.
  The outcome is random---sometimes it goes for the ore, sometimes for the tree.}
  \label{fig:dmd-ore-vs-tree}
\end{figure}

\begin{figure}[htbp]
\centering
  \centering
  \includegraphics[width=0.8\linewidth]{fig/attack-prob.png}
  
  \includegraphics[width=0.8\linewidth]{fig/max-ablation-impact-attack.png}
  
  \includegraphics[width=0.8\linewidth]{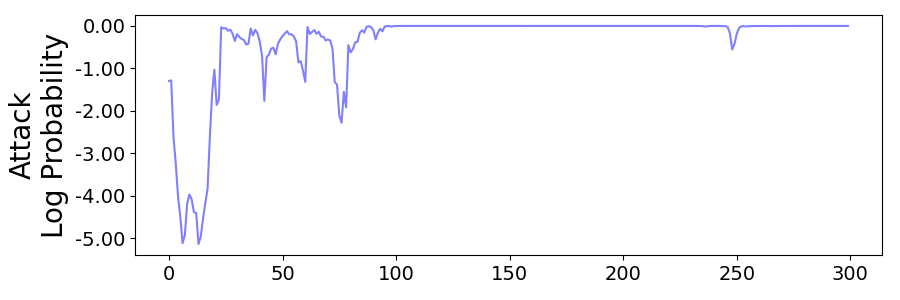}
  
  \includegraphics[width=0.8\linewidth]{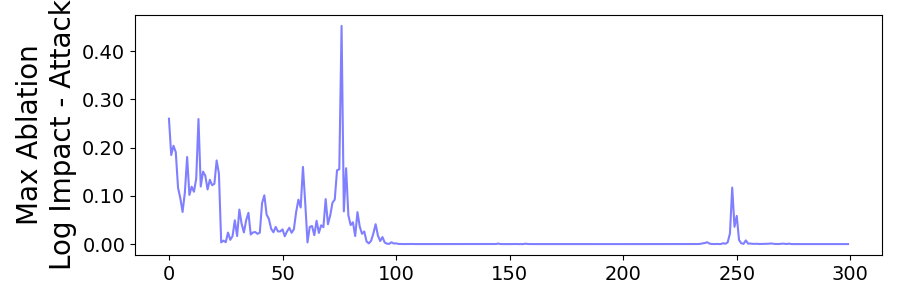}
  
  \hspace{4.5em}
  \includegraphics[width=.62\linewidth]{fig/episode-villager-partial.png}
\caption{
Comparison of measures of probabilities and log probabilities and their differences.
The top two plots show probabilities and max probability differences for ablations.
The bottom two plots show the same but in log probabilities and their max differences for ablations.
The only difference is that we can see some log probability ablation impact when the attack probability is near 0\% at the start of the episode.}
\label{fig:probabilities-vs-log-probabilities}
\end{figure}

\begin{figure}[htbp]
    \centering
    \begin{minipage}[t]{0.45\textwidth}
        \centering
        \includegraphics[width=\textwidth]{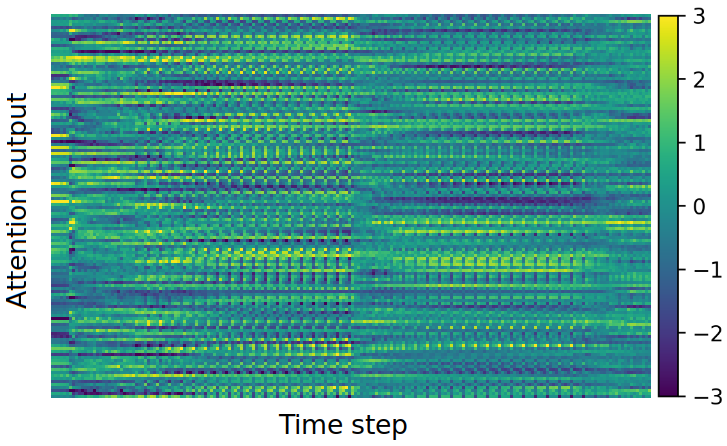} 
        \caption{Attention head 1.2 shows this unique pattern likely due to paying attention to every 4th frame, which can be seen in the attention weight plots (\autoref{fig:all-heads-layer-1}).}
        \label{fig:activation-viewer-1.2}
    \end{minipage}\hfill
    \begin{minipage}[t]{0.45\textwidth}
        \centering
        \includegraphics[width=\textwidth]{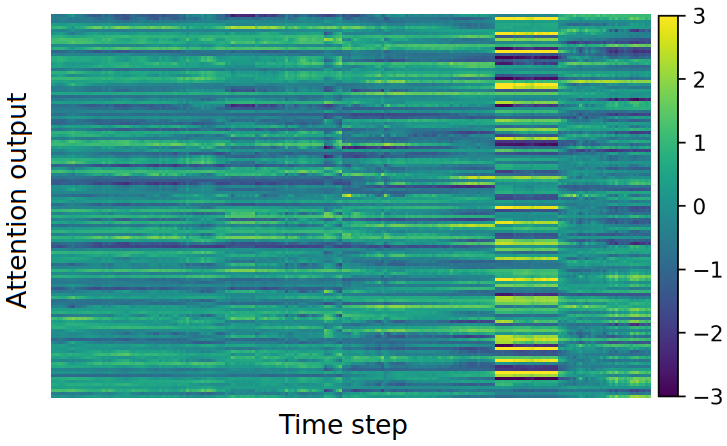} 
        \caption{
        Attention head 3.3 shows a distinct pattern (the one on the right) while the agent tries to open the crafting table menu.
        It starts right after the previous subtask is finished.
        The agent switches to the crafting table on the hotbar, jumps up, places it under itself, and opens it, after which the pattern stops.
        This pattern happens in multiple parts of the episode, whenever the crafting table menu is needed.
        See \href{https://youtu.be/uxghPuxh_0I}{Video13}.}
        \label{fig:activation-viewer-3.3}
    \end{minipage}
\end{figure}

\begin{figure}[htbp]
  \centering
  \includegraphics[width=\linewidth]{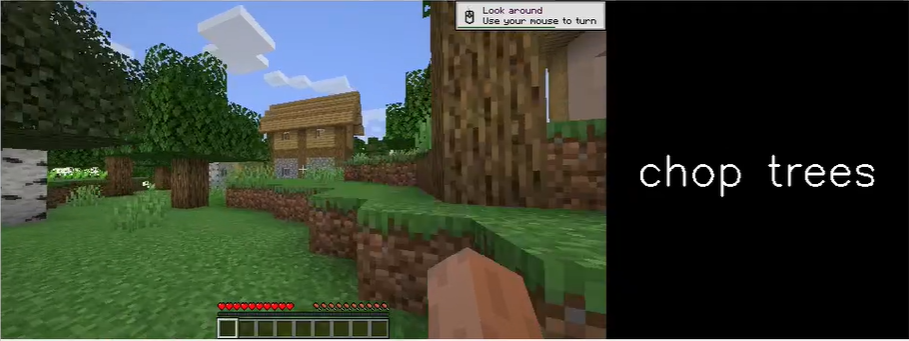}
  \caption{When given the command "chop trees" in this situation, Steve-1 runs towards the house and proceeds to spend several minutes destroying it.
  See \href{https://youtu.be/VQhP3h9nxqo}{Video14}.}
  \label{fig:steve-1-housechop}
\end{figure}

\begin{figure}[htbp]
  \centering
  \includegraphics[width=0.55\linewidth]{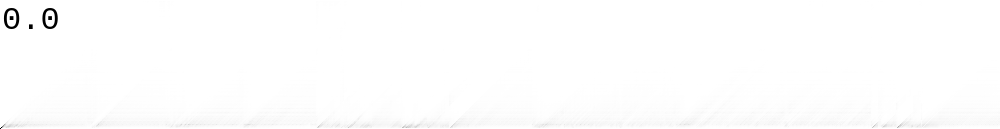}
  \includegraphics[width=0.55\linewidth]{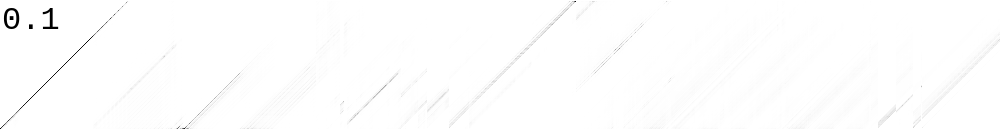}
  \includegraphics[width=0.55\linewidth]{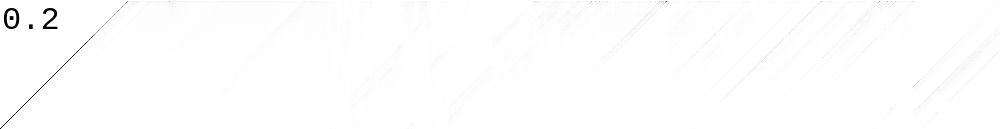}
  \includegraphics[width=0.55\linewidth]{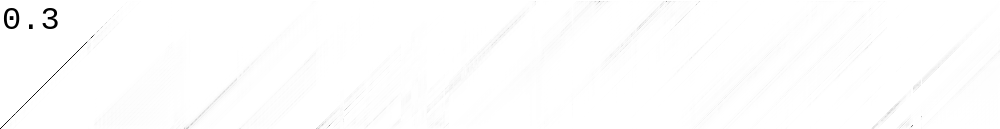}
  \includegraphics[width=0.55\linewidth]{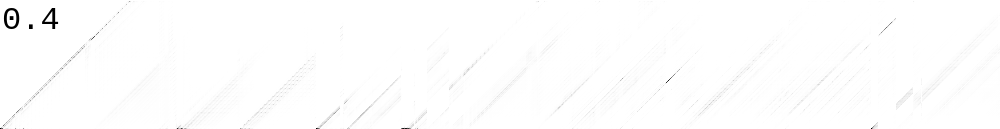}
  \includegraphics[width=0.55\linewidth]{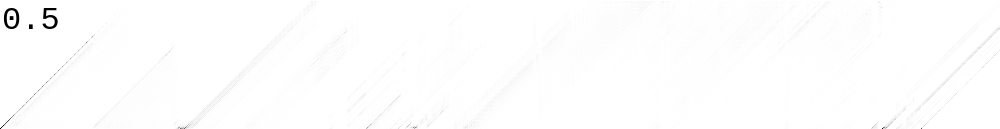}
  \includegraphics[width=0.55\linewidth]{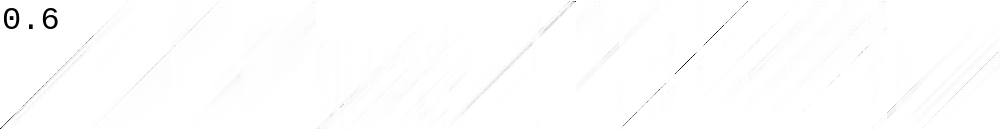}
  \includegraphics[width=0.55\linewidth]{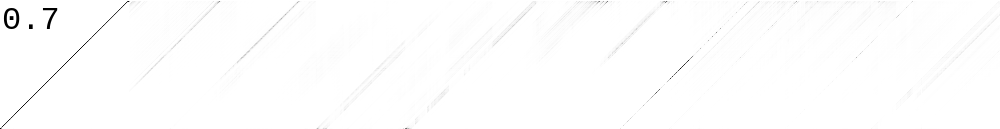}
  \includegraphics[width=0.55\linewidth]{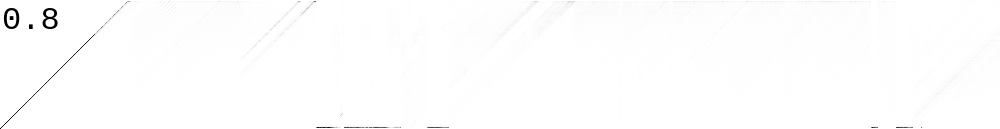}
  \includegraphics[width=0.55\linewidth]{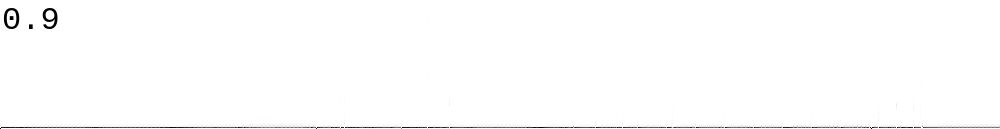}
  \includegraphics[width=0.55\linewidth]{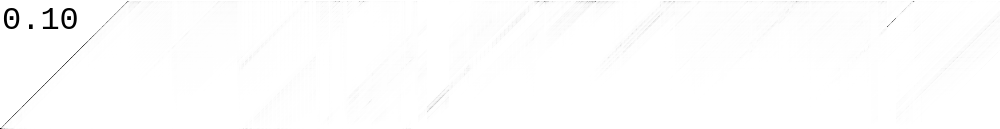}
  \includegraphics[width=0.55\linewidth]{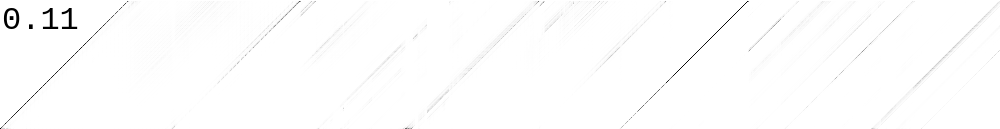}
  \includegraphics[width=0.55\linewidth]{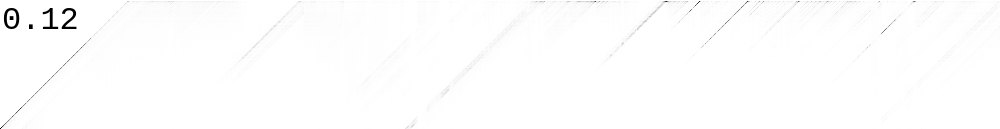}
  \includegraphics[width=0.55\linewidth]{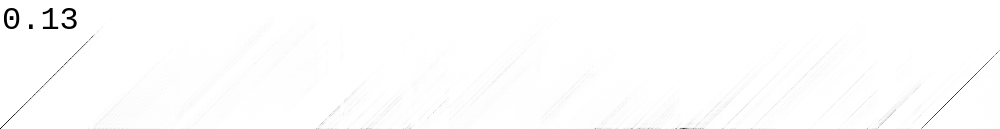}
  \includegraphics[width=0.55\linewidth]{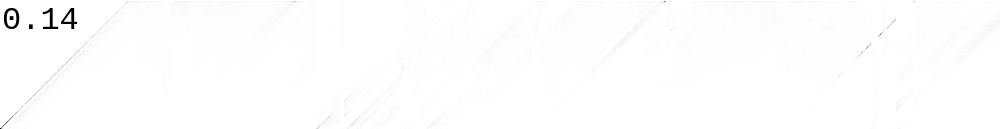}
  \includegraphics[width=0.55\linewidth]{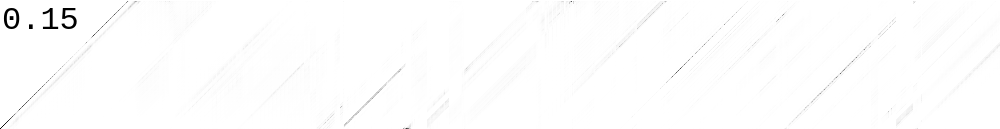}
  \includegraphics[width=0.55\linewidth]{fig/episode-stone-pick.png}
  \caption{Attention weights of attention heads in layer 0.}
  \label{fig:all-heads-layer-0}
\end{figure}

\begin{figure}[htbp]
  \centering
  \includegraphics[width=0.55\linewidth]{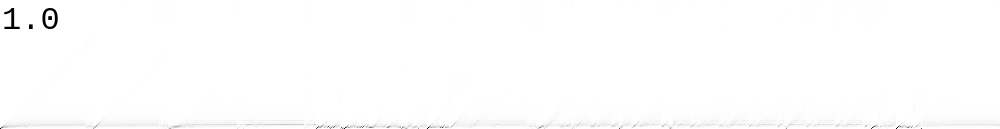}
  \includegraphics[width=0.55\linewidth]{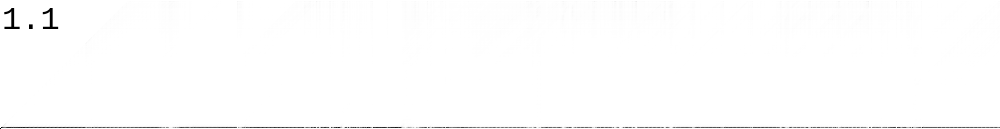}
  \includegraphics[width=0.55\linewidth]{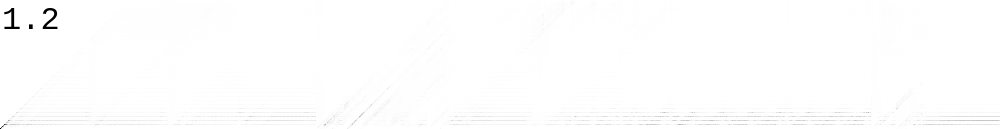}
  \includegraphics[width=0.55\linewidth]{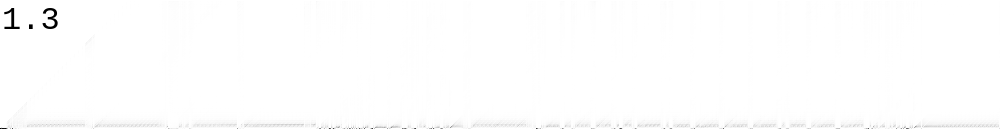}
  \includegraphics[width=0.55\linewidth]{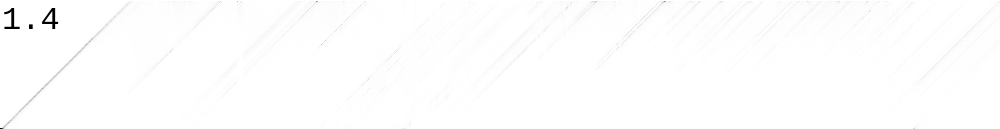}
  \includegraphics[width=0.55\linewidth]{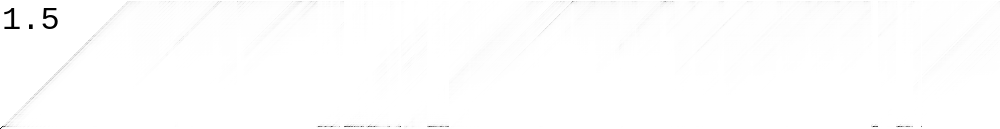}
  \includegraphics[width=0.55\linewidth]{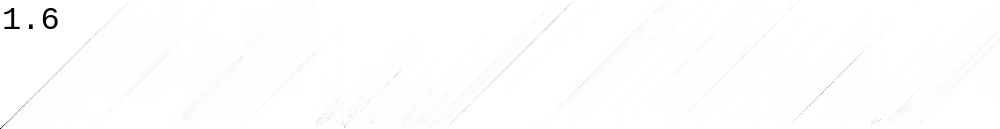}
  \includegraphics[width=0.55\linewidth]{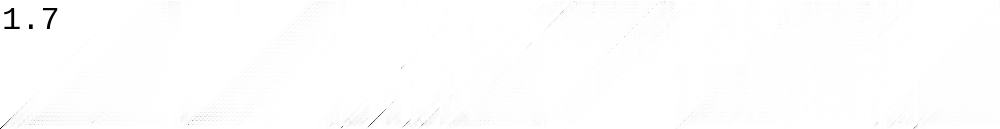}
  \includegraphics[width=0.55\linewidth]{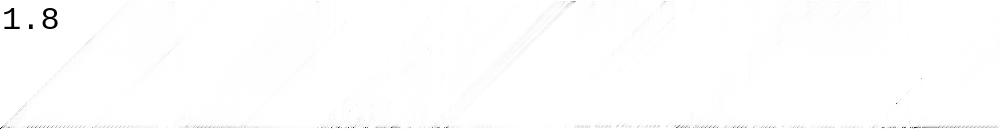}
  \includegraphics[width=0.55\linewidth]{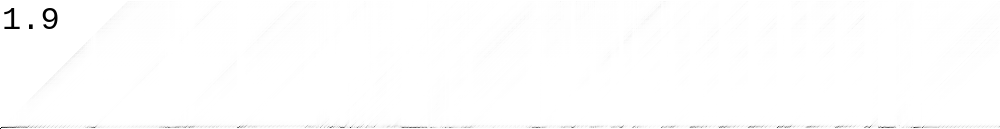}
  \includegraphics[width=0.55\linewidth]{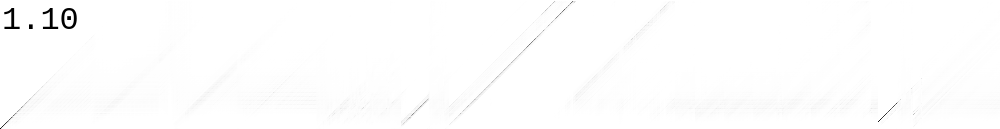}
  \includegraphics[width=0.55\linewidth]{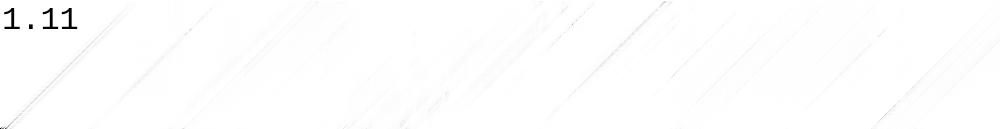}
  \includegraphics[width=0.55\linewidth]{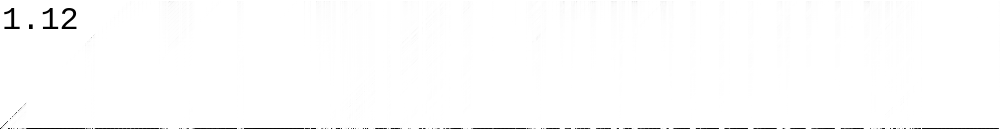}
  \includegraphics[width=0.55\linewidth]{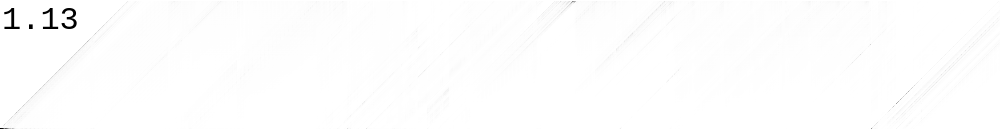}
  \includegraphics[width=0.55\linewidth]{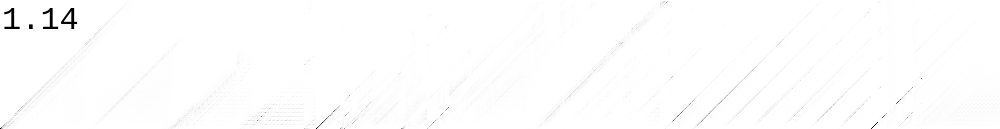}
  \includegraphics[width=0.55\linewidth]{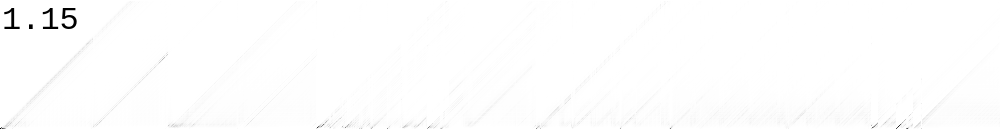}
  \includegraphics[width=0.55\linewidth]{fig/episode-stone-pick.png}
  \caption{Attention weights of attention heads in layer 1.}
  \label{fig:all-heads-layer-1}
\end{figure}

\begin{figure}[htbp]
  \centering
  \includegraphics[width=0.55\linewidth]{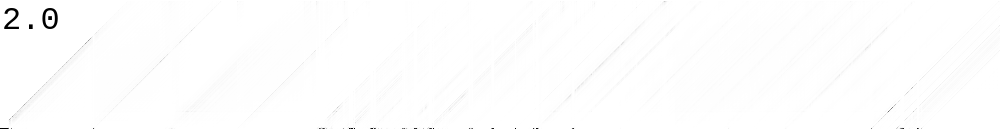}
  \includegraphics[width=0.55\linewidth]{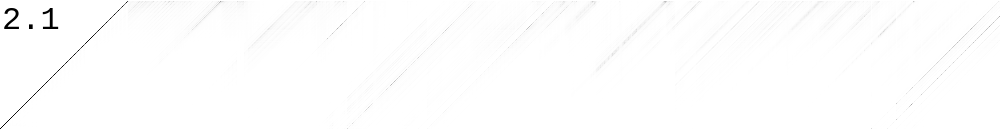}
  \includegraphics[width=0.55\linewidth]{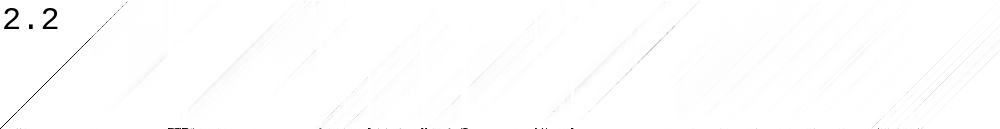}
  \includegraphics[width=0.55\linewidth]{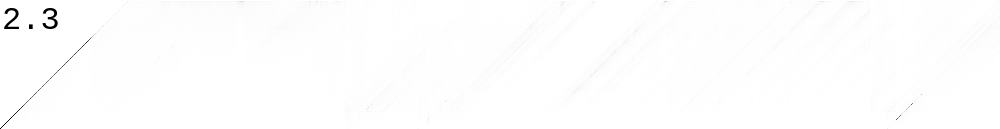}
  \includegraphics[width=0.55\linewidth]{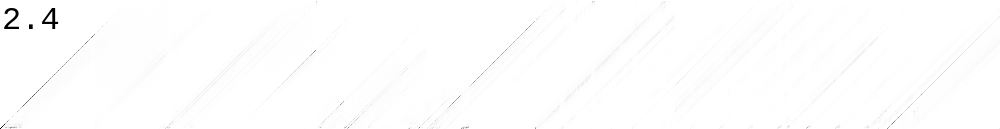}
  \includegraphics[width=0.55\linewidth]{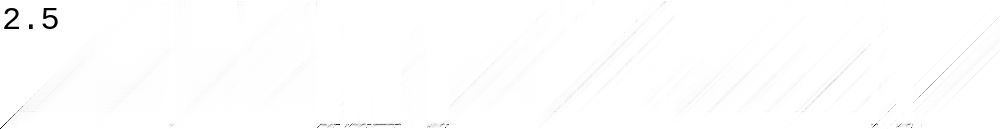}
  \includegraphics[width=0.55\linewidth]{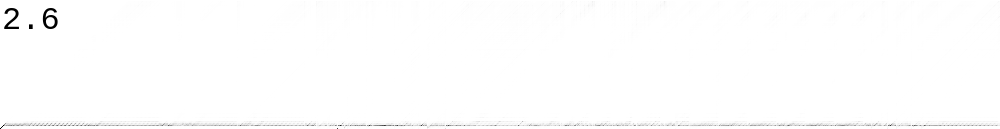}
  \includegraphics[width=0.55\linewidth]{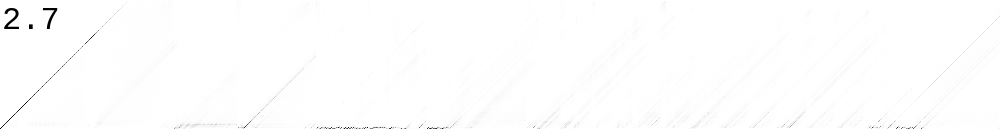}
  \includegraphics[width=0.55\linewidth]{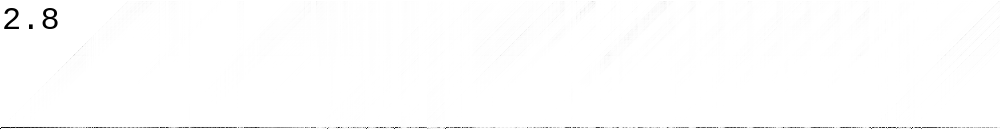}
  \includegraphics[width=0.55\linewidth]{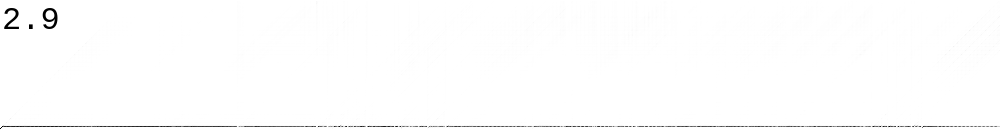}
  \includegraphics[width=0.55\linewidth]{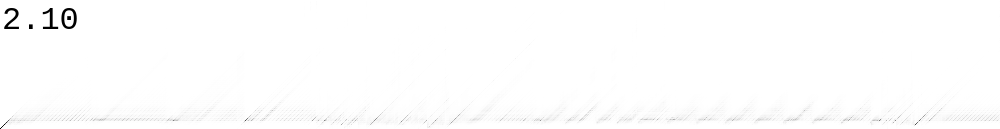}
  \includegraphics[width=0.55\linewidth]{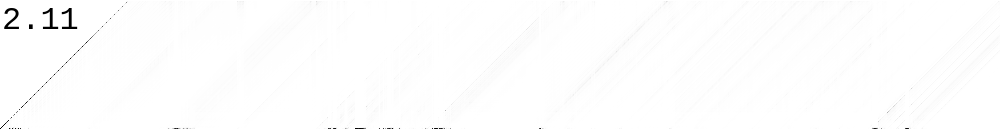}
  \includegraphics[width=0.55\linewidth]{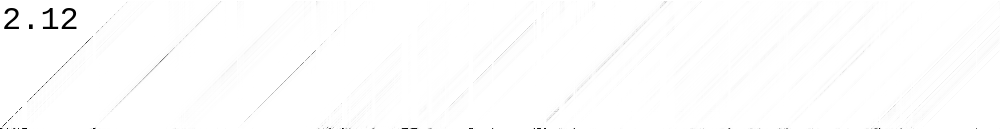}
  \includegraphics[width=0.55\linewidth]{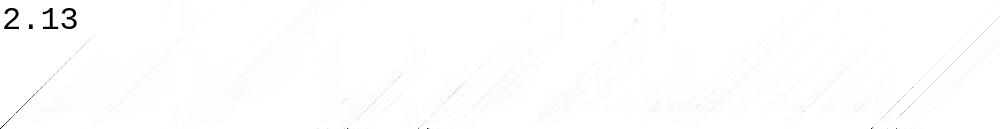}
  \includegraphics[width=0.55\linewidth]{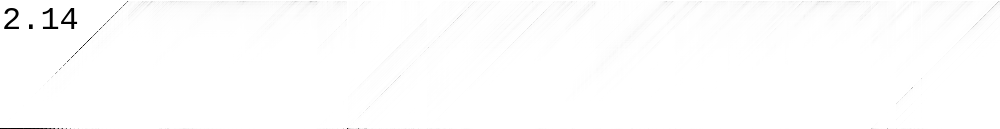}
  \includegraphics[width=0.55\linewidth]{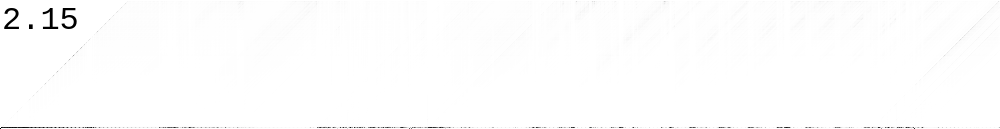}
  \includegraphics[width=0.55\linewidth]{fig/episode-stone-pick.png}
  \caption{Attention weights of attention heads in layer 2.}
  \label{fig:all-heads-layer-2}
\end{figure}

\begin{figure}[htbp]
  \centering
  \includegraphics[width=0.55\linewidth]{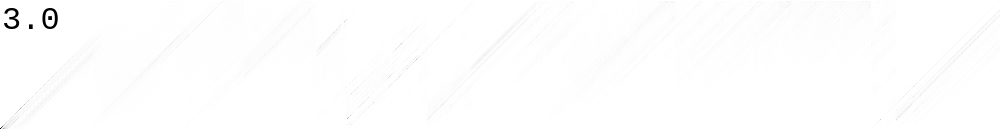}
  \includegraphics[width=0.55\linewidth]{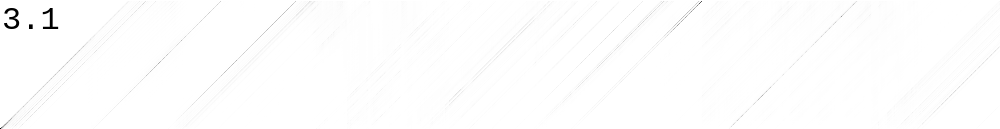}
  \includegraphics[width=0.55\linewidth]{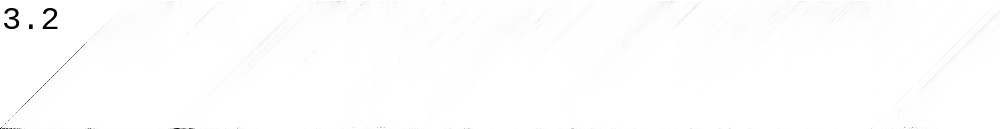}
  \includegraphics[width=0.55\linewidth]{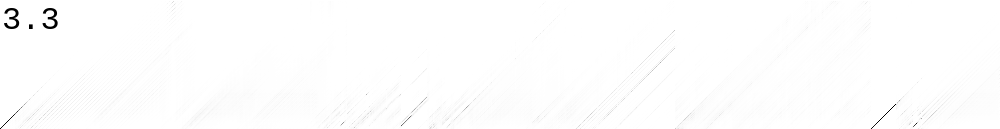}
  \includegraphics[width=0.55\linewidth]{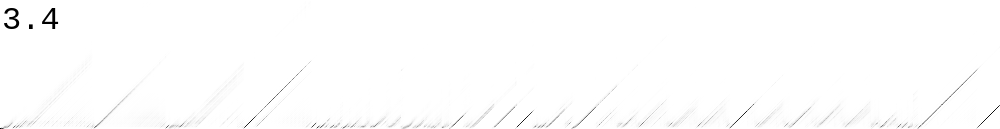}
  \includegraphics[width=0.55\linewidth]{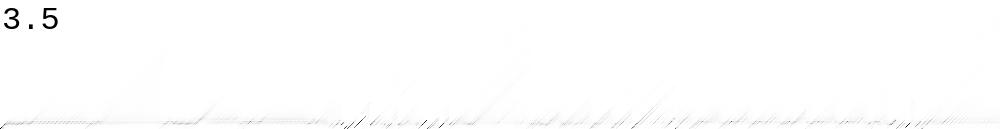}
  \includegraphics[width=0.55\linewidth]{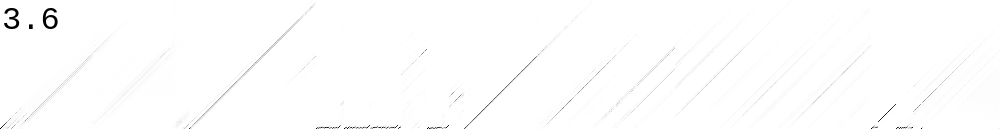}
  \includegraphics[width=0.55\linewidth]{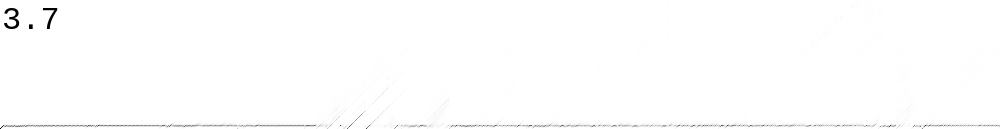}
  \includegraphics[width=0.55\linewidth]{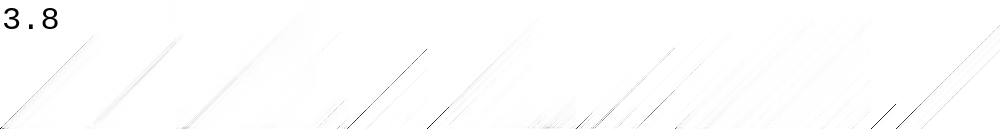}
  \includegraphics[width=0.55\linewidth]{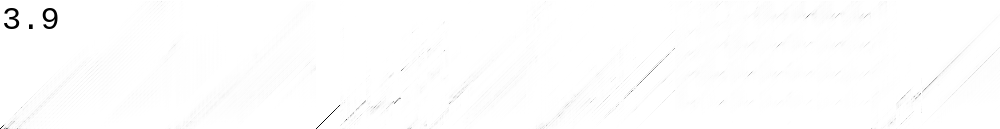}
  \includegraphics[width=0.55\linewidth]{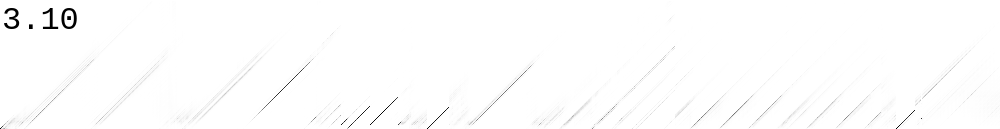}
  \includegraphics[width=0.55\linewidth]{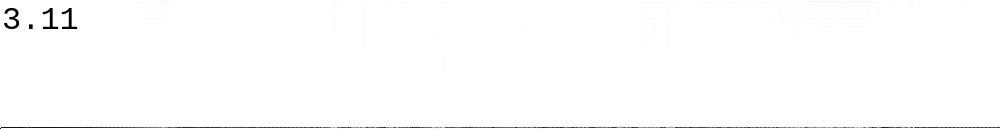}
  \includegraphics[width=0.55\linewidth]{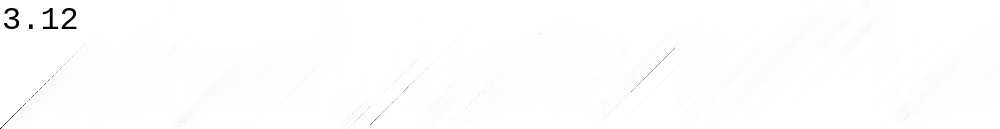}
  \includegraphics[width=0.55\linewidth]{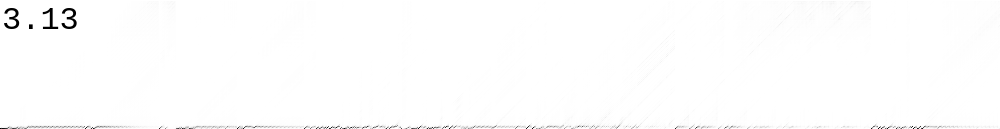}
  \includegraphics[width=0.55\linewidth]{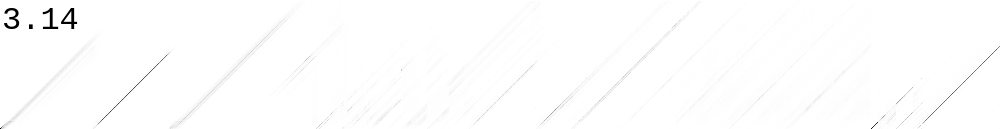}
  \includegraphics[width=0.55\linewidth]{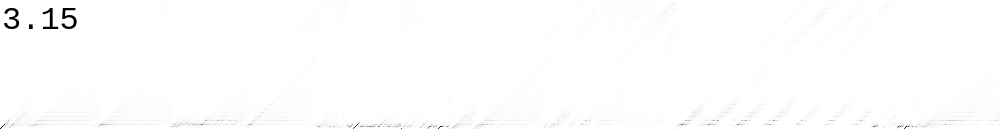}
  \includegraphics[width=0.55\linewidth]{fig/episode-stone-pick.png}
  \caption{Attention weights of attention heads in layer 3.}
  \label{fig:all-heads-layer-3}
\end{figure}

\clearpage
\newpage
\section{Videos}

A list of videos and their descriptions:

\begin{itemize}
    \item \href{https://youtu.be/g-jd6OyOcUs}{https://youtu.be/g-jd6OyOcUs} (Video01) Teaser.
    \item \href{https://youtu.be/BeqSthHRyLA}{https://youtu.be/BeqSthHRyLA} (Video02) Head 2.2 attention weights.
    \item \href{https://youtu.be/3GhhEysmSY4}{https://youtu.be/3GhhEysmSY4} (Video03) Frames with highest weights at each time step for each attention head.
    \item \href{https://youtu.be/TbTBWdb6jSo}{https://youtu.be/TbTBWdb6jSo} (Video04) Head 2.2 attention outputs.
    \item \href{https://youtu.be/e5qWNVEtuDA}{https://youtu.be/e5qWNVEtuDA} (Video05) VPT drops the four logs it was given at the start.
    \item \href{https://youtu.be/0-bxLngYO1Y}{https://youtu.be/0-bxLngYO1Y} (Video06) VPT with memory removed.
    \item \href{https://youtu.be/ju-s301cHzI}{https://youtu.be/ju-s301cHzI} (Video07) Single output ablation impact.
    \item \href{https://youtu.be/U8NYiudY5n8}{https://youtu.be/U8NYiudY5n8} (Video08) If attack probability below 99\%, don't attack.
    \item \href{https://youtu.be/cCRGOTRZQ8U}{https://youtu.be/cCRGOTRZQ8U} (Video09) Saliency maps.
    \item \href{https://youtu.be/i4RbOqFDlKc}{https://youtu.be/i4RbOqFDlKc} (Video10) Representation engineering. +3*(tree minus empty field).
    \item \href{https://youtu.be/8NcUdqmCY4k}{https://youtu.be/8NcUdqmCY4k} (Video11) Representation engineering. -3*(tree minus empty field).
    \item \href{https://youtu.be/VVkWWgwKf0M}{https://youtu.be/VVkWWgwKf0M} (Video12) MineRL VillagerChop-v0.
    \item \href{https://youtu.be/uxghPuxh_0I}{https://youtu.be/uxghPuxh\_0I} (Video13) Head 3.3 attention outputs.
    \item \href{https://youtu.be/VQhP3h9nxqo}{https://youtu.be/VQhP3h9nxqo} (Video14) STEVE-1 in a village, chopping "trees".
\end{itemize}

%% file: main.bbl
\begin{thebibliography}{46}
\providecommand{\natexlab}[1]{#1}
\providecommand{\url}[1]{\texttt{#1}}
\expandafter\ifx\csname urlstyle\endcsname\relax
  \providecommand{\doi}[1]{doi: #1}\else
  \providecommand{\doi}{doi: \begingroup \urlstyle{rm}\Url}\fi

\bibitem[Abi~Raad et~al.(2024)Abi~Raad, Ahuja, Barros, Besse, Bolt, Bolton, Brownfield, Buttimore, Cant, Chakera, Chan, Clune, Collister, Copeman, Cullum, Dasgupta, de~Cesare, Di~Trapani, Donchev, Dunleavy, Engelcke, Faulkner, Garcia, Gbadamosi, Gong, Gonzales, Gregor, Gupta, Hallingstad, Harley, Haves, Hill, Hirst, Hudson, Hudson, Hughes-Fitt, Rezende, Jasarevic, Kampis, Ke, Keck, Kim, Knagg, Kopparapu, Lampinen, Legg, Lerchner, Limont, Liu, Loks-Thompson, Marino, Martin~Cussons, Matthey, Mcloughlin, Mendolicchio, Merzic, Mitenkova, Moufarek, Oliveira, Oliveira, Openshaw, Pan, Pappu, Platonov, Purkiss, Reichert, Reid, Richemond, Roberts, Ruscoe, Sanchez~Elias, Sandars, Sawyer, Scholtes, Simmons, Slater, Soyer, Strathmann, Stys, Tam, Teplyashin, Terzi, Vercelli, Vujatovic, Wainwright, Wang, Wang, Wierstra, Williams, Wong, York, and Young]{raad2024scaling}
Abi~Raad, M., Ahuja, A., Barros, C., Besse, F., Bolt, A., Bolton, A., Brownfield, B., Buttimore, G., Cant, M., Chakera, S., Chan, S. C.~Y., Clune, J., Collister, A., Copeman, V., Cullum, A., Dasgupta, I., de~Cesare, D., Di~Trapani, J., Donchev, Y., Dunleavy, E., Engelcke, M., Faulkner, R., Garcia, F., Gbadamosi, C., Gong, Z., Gonzales, L., Gregor, K., Gupta, K., Hallingstad, A.~O., Harley, T., Haves, S., Hill, F., Hirst, E., Hudson, D.~A., Hudson, J., Hughes-Fitt, S., Rezende, D.~J., Jasarevic, M., Kampis, L., Ke, R., Keck, T., Kim, J., Knagg, O., Kopparapu, K., Lampinen, A., Legg, S., Lerchner, A., Limont, M., Liu, Y., Loks-Thompson, M., Marino, J., Martin~Cussons, K., Matthey, L., Mcloughlin, S., Mendolicchio, P., Merzic, H., Mitenkova, A., Moufarek, A., Oliveira, V., Oliveira, Y., Openshaw, H., Pan, R., Pappu, A., Platonov, A., Purkiss, O., Reichert, D., Reid, J., Richemond, P.~H., Roberts, T., Ruscoe, G., Sanchez~Elias, J., Sandars, T., Sawyer, D.~P., Scholtes, T., Simmons, G., Slater, D., Soyer, H.,
  Strathmann, H., Stys, P., Tam, A.~C., Teplyashin, D., Terzi, T., Vercelli, D., Vujatovic, B., Wainwright, M., Wang, J.~X., Wang, Z., Wierstra, D., Williams, D., Wong, N., York, S., and Young, N.
\newblock Scaling instructable agents across many simulated worlds.
\newblock \emph{arXiv preprint arXiv:2404.10179}, 2024.
\newblock URL \url{https://arxiv.org/abs/2404.10179}.

\bibitem[Adebayo et~al.(2018)Adebayo, Gilmer, Muelly, Goodfellow, Hardt, and Kim]{adebayo2018sanity}
Adebayo, J., Gilmer, J., Muelly, M., Goodfellow, I., Hardt, M., and Kim, B.
\newblock Sanity checks for saliency maps.
\newblock In \emph{NeurIPS 2018}, 2018.
\newblock URL \url{https://papers.nips.cc/paper/8160-sanity-checks-for-saliency-maps}.

\bibitem[Baker et~al.(2022)Baker, Akkaya, Zhokhov, Huizinga, Tang, Ecoffet, Houghton, Sampedro, and Clune]{baker2022video}
Baker, B., Akkaya, I., Zhokhov, P., Huizinga, J., Tang, J., Ecoffet, A., Houghton, B., Sampedro, R., and Clune, J.
\newblock Video pretraining (vpt): Learning to act by watching unlabeled online videos.
\newblock In \emph{NeurIPS 2022}, 2022.
\newblock URL \url{https://arxiv.org/abs/2206.11795}.

\bibitem[Barto \& Mahadevan(2003)Barto and Mahadevan]{barto2003hrl}
Barto, A.~G. and Mahadevan, S.
\newblock Recent advances in hierarchical reinforcement learning.
\newblock \emph{Discrete Event Dynamic Systems}, 13\penalty0 (4):\penalty0 341--379, 2003.
\newblock \doi{10.1023/A:1025696116075}.
\newblock URL \url{https://doi.org/10.1023/A:1025696116075}.

\bibitem[Beechey et~al.(2023)Beechey, Smith, and {\c{S}}im{\c{s}}ek]{beechey2023explaining}
Beechey, D., Smith, T.~M., and {\c{S}}im{\c{s}}ek, {\"O}.
\newblock Explaining reinforcement learning with shapley values.
\newblock In \emph{International Conference on Machine Learning}, pp.\  2003--2014. PMLR, 2023.

\bibitem[Chughtai et~al.(2023)Chughtai, Chan, and Nanda]{chughtai2023toy}
Chughtai, B., Chan, L., and Nanda, N.
\newblock A toy model of universality: Reverse engineering how networks learn group operations.
\newblock In \emph{International Conference on Machine Learning}, 2023.
\newblock URL \url{https://arxiv.org/abs/2302.03025}.

\bibitem[Conmy et~al.(2023)Conmy, Mavor-Parker, Lynch, Heimersheim, and Garriga-Alonso]{conmy2023towards}
Conmy, A., Mavor-Parker, A., Lynch, A., Heimersheim, S., and Garriga-Alonso, A.
\newblock Towards automated circuit discovery for mechanistic interpretability.
\newblock \emph{Advances in Neural Information Processing Systems}, 36:\penalty0 16318--16352, 2023.

\bibitem[Dao et~al.(2018)Dao, Mishra, and Lee]{dao2018deep}
Dao, G., Mishra, I., and Lee, M.
\newblock Deep reinforcement learning monitor for snapshot recording.
\newblock In \emph{Proceedings of the 2018 17th IEEE International Conference on Machine Learning and Applications (ICMLA ’18)}, pp.\  591--598, Los Alamitos, CA, 2018. IEEE.

\bibitem[Di~Langosco et~al.(2022)Di~Langosco, Koch, Sharkey, Pfau, and Krueger]{di2022goal}
Di~Langosco, L.~L., Koch, J., Sharkey, L.~D., Pfau, J., and Krueger, D.
\newblock Goal misgeneralization in deep reinforcement learning.
\newblock In \emph{International Conference on Machine Learning}, pp.\  12004--12019. PMLR, 2022.

\bibitem[D’Amour et~al.(2022)D’Amour, Heller, Moldovan, Adlam, Alipanahi, Beutel, Chen, Deaton, Eisenstein, Hoffman, et~al.]{damour2022underspecification}
D’Amour, A., Heller, K., Moldovan, D., Adlam, B., Alipanahi, B., Beutel, A., Chen, C., Deaton, J., Eisenstein, J., Hoffman, M.~D., et~al.
\newblock Underspecification presents challenges for credibility in modern machine learning.
\newblock \emph{Journal of Machine Learning Research}, 23:\penalty0 1--61, 2022.
\newblock URL \url{https://www.jmlr.org/papers/volume23/20-1335/20-1335.pdf}.
\newblock Submitted 11/20; Revised 12/21; Published 08/22.

\bibitem[Elhage et~al.(2021)Elhage, Nanda, Olsson, Henighan, Joseph, Mann, Askell, Bai, Chen, Conerly, et~al.]{elhage2021mathematical}
Elhage, N., Nanda, N., Olsson, C., Henighan, T., Joseph, N., Mann, B., Askell, A., Bai, Y., Chen, A., Conerly, T., et~al.
\newblock A mathematical framework for transformer circuits.
\newblock \emph{Transformer Circuits Thread}, 1:\penalty0 1, 2021.

\bibitem[{Figure}(2024)]{figure2024ai}
{Figure}.
\newblock Building ai powered robots, 2024.
\newblock URL \url{https://www.figure.ai/}.

\bibitem[Gandelsman et~al.(2024)Gandelsman, Efros, and Steinhardt]{gandelsman2024interpreting}
Gandelsman, Y., Efros, A.~A., and Steinhardt, J.
\newblock Interpreting clip’s image representation via text-based decomposition.
\newblock In \emph{International Conference on Learning Representations}, 2024.
\newblock URL \url{https://openreview.net/forum?id=5Ca9sSzuDp}.

\bibitem[Glanois et~al.(2021)Glanois, Weng, Zimmer, Li, Yang, Hao, and Liu]{glanois2021survey}
Glanois, C., Weng, P., Zimmer, M., Li, D., Yang, T., Hao, J., and Liu, W.
\newblock A survey on interpretable reinforcement learning.
\newblock \emph{arXiv preprint arXiv:2112.13112}, 2021.
\newblock URL \url{https://arxiv.org/pdf/2112.13112}.

\bibitem[Guss et~al.(2019)Guss, Houghton, Topin, Wang, Codel, Veloso, and Salakhutdinov]{gussminerl2020}
Guss, W.~H., Houghton, B., Topin, N., Wang, P., Codel, C., Veloso, M., and Salakhutdinov, R.
\newblock Minerl: {A} large-scale dataset of minecraft demonstrations.
\newblock \emph{CoRR}, abs/1907.13440, 2019.
\newblock URL \url{http://arxiv.org/abs/1907.13440}.

\bibitem[Gwynne \& Rentz(1983)Gwynne and Rentz]{gwynne1983beetles}
Gwynne, D.~T. and Rentz, D.~C.
\newblock Beetles on the bottle: male buprestids mistake stubbies for females (coleoptera).
\newblock \emph{Australian Journal of Entomology}, 22\penalty0 (1):\penalty0 79--80, 1983.

\bibitem[Hanna et~al.(2023)Hanna, Liu, and Variengien]{hanna2023does}
Hanna, M., Liu, O., and Variengien, A.
\newblock How does gpt-2 compute greater-than?: Interpreting mathematical abilities in a pre-trained language model.
\newblock In \emph{NeurIPS 2023}, 2023.

\bibitem[Heimersheim \& Nanda(2024)Heimersheim and Nanda]{heimersheim2024activation}
Heimersheim, S. and Nanda, N.
\newblock How to use and interpret activation patching.
\newblock \emph{arXiv preprint arXiv:2404.15255}, 2024.
\newblock URL \url{https://arxiv.org/abs/2404.15255}.

\bibitem[Hilton et~al.(2020)Hilton, Cammarata, Carter, Goh, and Olah]{hilton2020understanding}
Hilton, J., Cammarata, N., Carter, S., Goh, G., and Olah, C.
\newblock Understanding rl vision.
\newblock \emph{Distill}, 2020.
\newblock \doi{10.23915/distill.00029}.
\newblock https://distill.pub/2020/understanding-rl-vision.

\bibitem[Izmailov et~al.(2022)Izmailov, Kirichenko, Gruver, and Wilson]{izmailov2022feature}
Izmailov, P., Kirichenko, P., Gruver, N., and Wilson, A.~G.
\newblock On feature learning in the presence of spurious correlations.
\newblock \emph{Advances in Neural Information Processing Systems}, 35:\penalty0 38516--38532, 2022.

\bibitem[Joseph et~al.(2023)Joseph, Zholus, Samsami, and Richards]{joseph2023mining}
Joseph, S., Zholus, A., Samsami, M.~R., and Richards, B.~A.
\newblock Mining the diamond miner: Mechanistic interpretability on the video pretraining agent, 2023.
\newblock URL \url{https://openreview.net/pdf?id=lDeysxpH6W}.

\bibitem[Kenny et~al.(2023)Kenny, Tucker, and Shah]{kenny2023towards}
Kenny, E.~M., Tucker, M., and Shah, J.
\newblock Towards interpretable deep reinforcement learning with human-friendly prototypes.
\newblock In \emph{The Eleventh International Conference on Learning Representations}, 2023.
\newblock URL \url{https://openreview.net/forum?id=hWwY_Jq0xsN}.

\bibitem[Lieberum et~al.(2023)Lieberum, Rahtz, Kram{\'a}r, Irving, Shah, and Mikulik]{lieberum2023does}
Lieberum, T., Rahtz, M., Kram{\'a}r, J., Irving, G., Shah, R., and Mikulik, V.
\newblock Does circuit analysis interpretability scale? evidence from multiple choice capabilities in chinchilla.
\newblock \emph{arXiv preprint arXiv:2307.09458}, 2023.

\bibitem[Lifshitz et~al.(2023)Lifshitz, Paster, Chan, Ba, and McIlraith]{lifshitz2023steve1}
Lifshitz, S., Paster, K., Chan, H., Ba, J., and McIlraith, S.
\newblock Steve-1: A generative model for text-to-behavior in minecraft.
\newblock \emph{arXiv preprint arXiv:2306.00937}, 2023.

\bibitem[Linden(1999)]{Linden1999}
Linden, E.
\newblock \emph{The Parrot's Lament: And Other True Tales of Animal Intrigue, Intelligence, and Ingenuity}.
\newblock Dutton, New York, 1999.

\bibitem[Linden(2003)]{Linden2003}
Linden, E.
\newblock \emph{The Octopus and the Orangutan: New Tales of Animal Intrigue, Intelligence and Ingenuity}.
\newblock Penguin Publishing Group, New York, 2003.

\bibitem[McDougall et~al.(2023)McDougall, Conmy, Rushing, McGrath, and Nanda]{mcdougall2023copy}
McDougall, C., Conmy, A., Rushing, C., McGrath, T., and Nanda, N.
\newblock Copy suppression: Comprehensively understanding an attention head.
\newblock \emph{arXiv preprint arXiv:2310.04625}, 2023.

\bibitem[Milani et~al.(2023)Milani, Kanervisto, Ramanauskas, Schulhoff, Houghton, Mohanty, Galbraith, Chen, Song, Zhou, et~al.]{milani2023towards}
Milani, S., Kanervisto, A., Ramanauskas, K., Schulhoff, S., Houghton, B., Mohanty, S., Galbraith, B., Chen, K., Song, Y., Zhou, T., et~al.
\newblock Towards solving fuzzy tasks with human feedback: A retrospective of the minerl basalt 2022 competition.
\newblock \emph{arXiv preprint arXiv:2303.13512}, 2023.

\bibitem[Milani et~al.(2024)Milani, Topin, Veloso, and Fang]{milani2024explainable}
Milani, S., Topin, N., Veloso, M., and Fang, F.
\newblock Explainable reinforcement learning: A survey and comparative review.
\newblock \emph{ACM Comput. Surv.}, 56\penalty0 (7), apr 2024.
\newblock ISSN 0360-0300.
\newblock \doi{10.1145/3616864}.
\newblock URL \url{https://doi.org/10.1145/3616864}.

\bibitem[Mini et~al.(2023)Mini, Grietzer, Sharma, Meek, MacDiarmid, and Turner]{mini2023understanding}
Mini, U., Grietzer, P., Sharma, M., Meek, A., MacDiarmid, M., and Turner, A.~M.
\newblock Understanding and controlling a maze-solving policy network.
\newblock \emph{arXiv preprint arXiv:2310.08043}, 2023.

\bibitem[Mojang(2011)]{minecraft}
Mojang.
\newblock Minecraft.
\newblock \url{https://www.minecraft.net/}, 2011.

\bibitem[Olsson et~al.(2022{\natexlab{a}})Olsson, Elhage, Nanda, Joseph, DasSarma, Henighan, Mann, Askell, Bai, Chen, Conerly, Drain, Ganguli, Hatfield-Dodds, Hernandez, Johnston, Jones, Kernion, Lovitt, Ndousse, Amodei, Brown, Clark, Kaplan, McCandlish, and Olah]{olsson2022incontext}
Olsson, C., Elhage, N., Nanda, N., Joseph, N., DasSarma, N., Henighan, T., Mann, B., Askell, A., Bai, Y., Chen, A., Conerly, T., Drain, D., Ganguli, D., Hatfield-Dodds, Z., Hernandez, D., Johnston, S., Jones, A., Kernion, J., Lovitt, L., Ndousse, K., Amodei, D., Brown, T., Clark, J., Kaplan, J., McCandlish, S., and Olah, C.
\newblock In-context learning and induction heads, 2022{\natexlab{a}}.

\bibitem[Olsson et~al.(2022{\natexlab{b}})Olsson, Elhage, Nanda, Joseph, DasSarma, Henighan, Mann, Askell, Bai, Chen, et~al.]{olsson2022context}
Olsson, C., Elhage, N., Nanda, N., Joseph, N., DasSarma, N., Henighan, T., Mann, B., Askell, A., Bai, Y., Chen, A., et~al.
\newblock In-context learning and induction heads.
\newblock \emph{arXiv preprint arXiv:2209.11895}, 2022{\natexlab{b}}.

\bibitem[Puiutta \& Veith(2020)Puiutta and Veith]{puiutta2020explainable}
Puiutta, E. and Veith, E.~M.
\newblock Explainable reinforcement learning: A survey.
\newblock \emph{arXiv preprint arXiv:2005.06247}, 2020.

\bibitem[Ribeiro et~al.(2016)Ribeiro, Singh, and Guestrin]{ribeiro2016why}
Ribeiro, M.~T., Singh, S., and Guestrin, C.
\newblock "why should i trust you?": Explaining the predictions of any classifier.
\newblock In \emph{Proceedings of the 22nd ACM SIGKDD International Conference on Knowledge Discovery and Data Mining}, KDD '16, pp.\  1135–1144, New York, NY, USA, 2016. Association for Computing Machinery.
\newblock ISBN 9781450342322.
\newblock \doi{10.1145/2939672.2939778}.
\newblock URL \url{https://doi.org/10.1145/2939672.2939778}.

\bibitem[Sacks(2007)]{sacks2007musicophilia}
Sacks, O.
\newblock \emph{Musicophilia: Tales of Music and the Brain}.
\newblock Knopf, New York, 2007.

\bibitem[Sellam et~al.(2021)Sellam, Yadlowsky, Wei, Saphra, D'Amour, Linzen, Bastings, Turc, Eisenstein, Das, Tenney, and Pavlick]{sellam2021multiberts}
Sellam, T., Yadlowsky, S., Wei, J., Saphra, N., D'Amour, A., Linzen, T., Bastings, J., Turc, I., Eisenstein, J., Das, D., Tenney, I., and Pavlick, E.
\newblock The multiberts: Bert reproductions for robustness analysis.
\newblock \emph{ICLR'22}, 2021.
\newblock URL \url{https://arxiv.org/abs/2106.16163}.

\bibitem[Shah et~al.(2022)Shah, Varma, Kumar, Phuong, Krakovna, Uesato, and Kenton]{shah2022goal}
Shah, R., Varma, V., Kumar, R., Phuong, M., Krakovna, V., Uesato, J., and Kenton, Z.
\newblock Goal misgeneralization: Why correct specifications aren't enough for correct goals, 2022.

\bibitem[Simonyan et~al.(2014)Simonyan, Vedaldi, and Zisserman]{simonyan2014deep}
Simonyan, K., Vedaldi, A., and Zisserman, A.
\newblock Deep inside convolutional networks: Visualising image classification models and saliency maps, 2014.

\bibitem[Smilkov et~al.(2017)Smilkov, Thorat, Kim, Viégas, and Wattenberg]{smilkov2017smoothgrad}
Smilkov, D., Thorat, N., Kim, B., Viégas, F., and Wattenberg, M.
\newblock Smoothgrad: removing noise by adding noise.
\newblock \emph{arXiv preprint arXiv:1706.03825}, 2017.
\newblock URL \url{https://arxiv.org/abs/1706.03825}.

\bibitem[Templeton et~al.(2024)Templeton, Conerly, Marcus, Lindsey, Bricken, Chen, Pearce, Citro, Ameisen, Jones, Cunningham, Turner, McDougall, MacDiarmid, Freeman, Sumers, Rees, Batson, Jermyn, Carter, Olah, and Henighan]{templeton2024scaling}
Templeton, A., Conerly, T., Marcus, J., Lindsey, J., Bricken, T., Chen, B., Pearce, A., Citro, C., Ameisen, E., Jones, A., Cunningham, H., Turner, N.~L., McDougall, C., MacDiarmid, M., Freeman, C.~D., Sumers, T.~R., Rees, E., Batson, J., Jermyn, A., Carter, S., Olah, C., and Henighan, T.
\newblock Scaling monosemanticity: Extracting interpretable features from claude 3 sonnet.
\newblock \emph{Transformer Circuits Thread}, 2024.
\newblock URL \url{https://transformer-circuits.pub/2024/scaling-monosemanticity/index.html}.

\bibitem[Turner et~al.(2024)Turner, Thiergart, Leech, Udell, Vazquez, Mini, and MacDiarmid]{turner2024activation}
Turner, A.~M., Thiergart, L., Leech, G., Udell, D., Vazquez, J.~J., Mini, U., and MacDiarmid, M.
\newblock Activation addition: Steering language models without optimization, 2024.

\bibitem[Wang et~al.(2022)Wang, Variengien, Conmy, Shlegeris, and Steinhardt]{wang2022interpretability}
Wang, K., Variengien, A., Conmy, A., Shlegeris, B., and Steinhardt, J.
\newblock Interpretability in the wild: a circuit for indirect object identification in gpt-2 small.
\newblock \emph{arXiv preprint arXiv:2211.00593}, 2022.

\bibitem[Xu \& Fekri(2021)Xu and Fekri]{xu2021interpretable}
Xu, D. and Fekri, F.
\newblock Interpretable model-based hierarchical reinforcement learning using inductive logic programming.
\newblock \emph{CoRR}, abs/2106.11417, 2021.
\newblock URL \url{https://arxiv.org/abs/2106.11417}.

\bibitem[Zolman et~al.(2024)Zolman, Fasel, Kutz, and Brunton]{zolman2024sindyrl}
Zolman, N., Fasel, U., Kutz, J.~N., and Brunton, S.~L.
\newblock Sindy-rl: Interpretable and efficient model-based reinforcement learning.
\newblock \emph{arXiv preprint arXiv:2403.09110}, 2024.

\bibitem[Zou et~al.(2023)Zou, Phan, Chen, Campbell, Guo, Ren, Pan, Yin, Mazeika, Dombrowski, Goel, Li, Byun, Wang, Mallen, Basart, Koyejo, Song, Fredrikson, Kolter, and Hendrycks]{zou2023representation}
Zou, A., Phan, L., Chen, S., Campbell, J., Guo, P., Ren, R., Pan, A., Yin, X., Mazeika, M., Dombrowski, A.-K., Goel, S., Li, N., Byun, M.~J., Wang, Z., Mallen, A., Basart, S., Koyejo, S., Song, D., Fredrikson, M., Kolter, J.~Z., and Hendrycks, D.
\newblock Representation engineering: A top-down approach to ai transparency, 2023.

\end{thebibliography}
